# Leveraging generative AI for human understanding: meta-requirements and design principles for Explanatory AI as an emerging paradigm

Christian Meske[a], Justin Brenne[a], Erdi Ünal[a], Sabahat Ölcer[b,c] and Aysegül Dogangün[b]

[a]Faculties of Mechanical Engineering and Computer Science, Institute of Work Science, Ruhr-Universität Bochum, Bochum, Germany; [b]Department of Human-Centered Technology Development, Institute of Computer Science, Ruhr West University of Applied Sciences, Bottrop, Germany; [c]Department of Palliative Medicine, University Medical Center Göttingen, University of Göttingen, Göttingen, Germany

**ABSTRACT**

Artificial intelligence (AI) systems increasingly support decision-making across critical domains, yet explainable AI (XAI) approaches prioritise algorithmic transparency over human comprehension. While XAI reveals computational processes for model validation, end users require explanations integrating domain knowledge, contextual reasoning, and professional frameworks. This disconnect reveals a design challenge: existing approaches fail to address how practitioners need to understand and act upon AI recommendations. This paper introduces Explanatory AI as an emerging complementary paradigm where AI systems leverage generative and multimodal capabilities to serve as explanatory partners for human understanding. Unlike traditional XAI, which answers 'How did the algorithm decide?' for validation purposes, Explanatory AI addresses 'Why does this make sense?' for practitioners making informed decisions. Through theory-informed design, we synthesise multidisciplinary perspectives from cognitive science, communication research, and education with empirical evidence from healthcare and AI expert interviews. Our analysis identifies five dimensions distinguishing Explanatory AI from XAI: explanatory purpose, communication mode, epistemic stance, adaptivity, and cognitive design. We derive five meta-requirements and formulate ten design principles prescribing how to build such systems, illustrated through five UI mockups exemplifying implementation. Explanatory AI operates standalone or combined with XAI, serving distinct purposes in creating systems that genuinely support human reasoning and decision-making.

## 1. Introduction

Artificial intelligence (AI) systems increasingly provide decision support across critical domains, yet a fundamental gap exists between how these systems explain their decisions and what users actually need to understand and act upon recommendations. Consider a healthcare scenario where an AI system recommends medication dosage adjustments: current explainable AI (XAI) approaches typically present feature attribution scores (e.g. SHAP values showing age +0.12, blood pressure −0.08, kidney function +0.15) that reveal algorithmic decision processes (Arrieta et al. 2020; Lundberg and Lee 2017). However, practitioners require explanations addressing why the recommendation makes clinical sense, understanding that integrates medical reasoning, patient context, and established care protocols. This disconnect reveals a critical design challenge: existing AI explanation approaches prioritise algorithmic transparency over human comprehension (Bove et al. 2024; Miller 2019). A growing body of work on Human-Centred AI highlights that genuine alignment with human values and needs requires moving beyond technical transparency towards designs that support comprehension, oversight, and human welfare (Schmager, Pappas, and Vassilakopoulou 2025).

Current XAI methods, including post-hoc techniques like SHAP and LIME (Lundberg and Lee 2017; Ribeiro, Singh, and Guestrin 2016) provide algorithmic transparency and are widely used for model validation, debugging, regulatory compliance, and algorithmic auditing (Adadi and Berrada 2018; Meske et al. 2022). However, end users who must understand, trust, and act upon AI recommendations face different challenges: they need explanations that help them make sense of recommendations within their domain expertise and situational contexts (Miller 2019; Shin 2021). Research

**CONTACT** Christian Meske christian.meske@rub.de Faculties of Mechanical Engineering and Computer Science, Institute of Work Science, Ruhr-Universität Bochum, Universitätsstr. 150, 44801, Bochum, Germany indicates that feature importance scores and statistical attributions, while technically accurate, leave (non-expert) users seeking fundamentally different types of understanding to support their decision-making processes (Bove et al. 2024; Chromik and Schuessler 2020; Kaur et al. 2020; Kim et al. 2024).

The emergence of generative artificial intelligence has created new possibilities for explanation that extend beyond algorithmic transparency towards human-centered communication. Large language models (LLMs) can produce contextual, narrative-driven explanations that adapt to user backgrounds, embed decisions within familiar conceptual frameworks, and support interactive dialogue (Chen et al. 2024; Martens et al. 2025; Singh et al. 2024; Slack et al. 2023). This capability opens opportunities for AI systems designed to help humans understand complex phenomena and make informed decisions (Raees et al. 2024; Singh et al. 2023, 2024).

We term this emerging approach ‘Explanatory AI’, defined functionally by its explanatory purpose: enabling human understanding and informed decision-making through contextual, narrative-driven explanations that integrate domain knowledge, adapt to user characteristics, and prioritise comprehension over algorithmic transparency. Generative AI capabilities currently provide the most viable technical means of achieving this purpose, but Explanatory AI denotes the purpose and design objective rather than any specific technology. Explanatory AI complements rather than replaces XAI, operating standalone or in combination with XAI methods. Unlike XAI methods that reveal computational processes for model validation, Explanatory AI prioritises user comprehension by integrating domain knowledge, employing dynamic narrative communication, and adapting to diverse user populations. Because Explanatory AI systems shape how practitioners understand and act on AI recommendations in consequential settings, their design also raises governance and accountability questions: for instance, how to ensure explanations remain faithful and domain-valid, and how responsibility is allocated when comprehensible but imperfect explanations inform high-stakes decisions. We return to these implications in Section 5. However, realising Explanatory AI’s potential requires systematic design guidance. The field lacks systematic, empirically grounded meta-requirements and design principles for creating such Explanatory AI systems. This gap raises two guiding research questions:

> RQ1: What is Explanatory AI, and what dimensions distinguish it from traditional XAI?
>
> RQ2: What meta-requirements and design principles should guide the development of Explanatory AI systems for high-stakes professional contexts?

We address these questions through theory-informed design (Gregor 2006; Gregor, Chandra Kruse, and Seidel 2020), synthesising insights from multidisciplinary explanation theory with empirical evidence from healthcare contexts to derive actionable design principles for Explanatory AI. Our approach combines theoretical analysis of explanation across cognitive science, communication research, and education with empirical meta-requirements gathering through Rapid Contextual Design methodology in healthcare settings and expert interviews with AI researchers and practitioners.

Our conceptualisation identifies five key dimensions that distinguish Explanatory AI from traditional XAI approaches: (1) explanatory purpose, shifting from diagnostic transparency to interpretive sense-making; (2) communication mode, moving from static technical outputs to dynamic narrative interaction; (3) epistemic stance, prioritising contextual plausibility over exhaustive algorithmic detail where appropriate; (4) adaptivity, enabling personalised universal accessibility; and (5) cognitive design, managing information flow for comprehension.

Our contributions are threefold. First, we conceptualise Explanatory AI as an emerging, complementary design paradigm to XAI, establishing a theoretical framework identifying five key dimensions (RQ1). Second, we provide empirical evidence from healthcare professionals and AI experts supporting this conceptualisation and derive five meta-requirements specifying what Explanatory AI systems must achieve (RQ1/RQ2). Third, we formulate ten design principles prescribing how to build systems that satisfy these meta-requirements, offering actionable guidance for practitioners (RQ2).

The remainder of this paper proceeds as follows. Section 2 describes our research approach, integrating theory-informed design with empirical investigation through contextual interviews with healthcare professionals and AI experts. Section 3 develops our conceptualisation by examining multidisciplinary perspectives on explanation with empirical evidence to derive five design dimensions and corresponding meta-requirements. Section 4 translates meta-requirements into ten actionable design principles for building Explanatory AI systems. Section 5 discusses theoretical and practical implications alongside limitations. Section 6 concludes with a summary of contributions and directions for future research.

## 2. Research approach

Drawing on design-theory foundations (Gregor 2006; Gregor, Chandra Kruse, and Seidel 2020), we adopt a theory-informed design approach to derive meta-requirements and design principles for Explanatory AI systems that prioritise human understanding over algorithmic transparency. This approach systematically integrates theoretical foundations with empirical evidence to generate actionable design knowledge in the form of meta-requirements and design principles. Our approach identifies five theoretical dimensions characterising the design space for Explanatory AI systems, examines these dimensions against empirical evidence in healthcare contexts and with AI experts, derives meta-requirements specifying what systems must achieve for each dimension, and formulates corresponding design principles prescribing how to achieve those requirements. Additionally we present five UI Mockups as exemplary implementations of the design principles. Theory-informed design recognises that effective Explanatory AI requires both grounding in established understanding of human cognition and communication, and real-world user needs. Our approach proceeds through three interconnected phases: theoretical analysis to identify design dimensions, empirical investigation to derive meta-requirements, and synthesis to formulate design principles.

In the *first phase*, we conducted integrative literature review (Torraco 2005) across two domains to establish theoretical foundations for our design dimensions: (1) multidisciplinary perspectives on human explanation, and (2) the evolution of AI explanation capabilities. This approach synthesises diverse literatures to generate new conceptual frameworks, which we applied by examining each domain in turn and then comparing them to derive design dimensions. We focus on these two domains because deriving design dimensions for Explanatory AI requires understanding both sides of the gap motivating this paper: what constitutes effective explanation for humans, and what current AI systems are technically capable of providing. First, we examined multidisciplinary perspectives on human explanation, drawing from philosophy of science, cognitive psychology, social psychology, linguistics, and education research. This analysis revealed recurring principles of effective human explanation: causal reasoning (Lombrozo 2012; Salmon 1984), contrastive framing (Hilton and Slugoski 1986; Miller 2019), audience adaptation (Grice 1975; Malle 2004), and narrative coherence (Dahlstrom 2014). Second, we traced the evolution of AI explanation capabilities from rule-based expert systems through statistical XAI methods to modern generative AI approaches. By comparing human explanation principles with AI explanation capabilities, we identified systematic design challenges where current approaches fail to meet human needs. This analysis yielded five design dimensions characterising the space of design decisions for Explanatory AI.

Consistent with Torraco's (2005) methodology, this integrative review is a purposive, conceptual synthesis rather than a systematic review: it aims to integrate perspectives across mature but disconnected literatures into a new framework, not to achieve exhaustive or reproducible coverage through a fixed search protocol. Accordingly, we did not apply a predefined keyword search string, database, or publication-date range, nor a formal screening or PRISMA-style inclusion/exclusion procedure. Instead, source selection was purposive and theory-anchored: we anchored our review of human explanation in seminal, widely-cited theoretical accounts spanning philosophy of science, cognitive and social psychology, linguistics, and education (e.g. Miller 2019, as a key synthesising reference for explanation in AI contexts), and traced the technical literature on AI explanation chronologically from early rule-based expert systems through statistical XAI methods to recent generative approaches. Coverage was thus governed by conceptual sufficiency rather than exhaustiveness: we did not aim to enumerate all types of human explanation, but to identify the principles that recur convergently across these literatures and are already consolidated in established synthesising accounts, and we stopped adding sources for a given principle once further accounts converged on principles already represented in the emerging dimension set. Selection decisions were made by the interdisciplinary author team and discussed informally with colleagues at the participating research chairs, approximately ten researchers spanning computer science, XAI, large language models and generative AI, information systems, positive computing, and the social sciences, whose feedback on candidate sources and on the framing of the dimensions informed successive rounds of revision. This was a formative, discussion-based process rather than a structured expert elicitation, and final selection responsibility rests with the author team. Triangulation across the two domains occurred at the synthesis stage (Section 3.1–3.2): we cross-checked whether principles identified in the human-explanation literature were structurally addressed, partially addressed, or unaddressed by current AI explanation capabilities, using convergence and divergence between the two literatures to derive the five design dimensions reported in Section 3.3. Because this synthesis was conceptual rather than data-coded, dimension formation followed an iterative, interpretive clustering process: recurring themes surfacing across both literatures were provisionally grouped by the author team, discussed against the convergence and divergence patterns identified through triangulation, and consolidated into five dimensions through repeated rounds of joint review.

Building on these theoretical foundations, in the *second phase*, we conducted qualitative fieldwork in healthcare contexts to understand how design dimensions manifest in practice and to derive meta-requirements. Our study focused on home care settings in Germany, exemplifying critical challenges: diverse end-user populations with varying technical literacy and language proficiency, high-stakes decisions, time-pressured environments, and needs spanning immediate task support and long-term learning. We employed Rapid Contextual Design (RCD) with six nursing staff members across German facilities (Holtzblatt, Wendell, and Wood 2004), purposefully selected to represent varying qualification levels and language backgrounds. Nursing staff were recruited through an ongoing research partnership with a home care institution in Germany. Given the applied, project-embedded nature of this collaboration, participant assignment was coordinated by the institution's care management according to diversity criteria specified by the research team, varying qualification levels (e.g. registered nurses, trainees, nursing assistants) and language backgrounds, reflecting the increasing linguistic diversity of the German care workforce. Interviews and observations took place during participants' regular working shifts, without additional compensation, minimising selection bias towards particularly engaged or available staff. Contextual interviews combined semi-structured dialogue with direct observation, documenting explanation needs during care tasks. We also conducted six semi-structured interviews with AI experts (XAI researchers, generative AI researchers, computer scientists, social science experts) to capture socio-technical stakeholder perspectives. AI experts were recruited via purposive (criterion) sampling from the authors' interdisciplinary professional network. Inclusion required demonstrated applied experience with AI or generative AI systems; participants were selected to span complementary disciplinary perspectives relevant to AI explanation, including computer science, psychology, and sociology, ensuring both technical and socio-behavioral viewpoints were represented among the six expert interviews. All interviews were conducted and recorded verbatim in German, transcribed using a locally deployed Python-based Whisper model, and relevant excerpts were subsequently translated into English for reporting purposes. Translations were generated using DeepL and subsequently cross-checked for accuracy by two additional authors fluent in both German and English, with discrepancies resolved through discussion. RCD analysis followed iterative affinity mapping (Beyer and Holtzblatt 1999), systematically clustering observations into themes organised by our five theoretical dimensions. Consistent with the method, clustering proceeded bottom-up: interview and observation notes were first grouped by substantive similarity on a shared digital whiteboard by four researchers, and the resulting clusters were only subsequently related to the five dimensions. Material that did not map onto a dimension was retained in a residual group rather than discarded, so that themes outside the framework could surface; the residual material concerned organisational and workflow conditions (e.g. staffing and shift constraints) that shape explanation needs without constituting a distinct explanatory design dimension, and no additional dimension emerged. For each dimension, we integrated theoretical tensions with empirical evidence to establish comprehensive meta-requirements for Explanatory AI (presented in Section 3.3). The study was approved by the Institutional Review Board (IRB) and conducted in full compliance with the Declaration of Helsinki. Informed consent was obtained from all participating elderly care nurses prior to data collection, which included interviews and observations.

AI expert interviews were analysed separately from the RCD nurse data using deductive coding (Hsieh and Shannon 2005): the five theoretical dimensions derived in the first phase (Section 3.1–3.2) served as an a priori coding framework applied to the interview transcripts. Coding was carried out by the same researchers involved in the affinity mapping, working on the same whiteboard, and included an explicit residual category for statements not attributable to any dimension. We did not develop a formal codebook, as the dimensions themselves, together with their definitions in Sections 3.1–3.2, served as the coding frame. Disagreements were resolved by discussion between the four coders; where this did not produce agreement, the wider author group involved was consulted to reach a joint decision. Within this framework, expert statements were clustered into recurring subthemes per dimension, following the same logic of integrating theoretical tensions with empirical evidence described above. For each dimension, the resulting expert-interview clusters were combined with the nursing RCD findings and the theoretical tension to formulate one meta-requirement (five total, Section 3.3), and each meta-requirement was subsequently decomposed into two design principles (ten total, Section 4).

In the *third phase*, we synthesised theoretical foundations and empirical evidence to derive meta-requirements and design principles for each dimension, consistent with Gregor's (2006) characterisation of design and action theory. This approach structures the derivation of prescriptive knowledge by integrating theoretical constructs with empirical evidence to formulate actionable meta-requirements and principles. For each dimension, we first formulated meta-requirements specifying what Explanatory AI systems must achieve, addressing the entire class of systems rather than specific implementations. We then derived design principles prescribing how to satisfy these meta-requirements (presented in Section 4). For each dimension, we integrated theoretical understanding of effective explanation (Lombrozo 2012; Miller 2019) with empirical evidence of user needs and current system failures. This synthesis process involved identifying where theory and empirics converged on specific meta-requirements, resolving tensions between different stakeholder needs, and formulating prescriptive principles offering actionable design guidance. Triangulation across multidisciplinary theory, healthcare user observations, and expert insights provided convergent support for derived meta-requirements and principles. The resulting meta-requirements and design principles represent theory-informed, empirically-grounded guidance for creating Explanatory AI systems that genuinely serve human understanding across diverse contexts and user populations.

Figure 1 summarises this chain of evidence. The integrative literature review (Phase 1, Section 3.1) yielded five design dimensions, which then structured both empirical strands of Phase 2: the RCD affinity mapping of nursing interviews and observations, where bottom-up clusters were subsequently related to the dimensions, and the deductive coding of AI expert interview transcripts, where the dimensions served as an a priori coding framework. Within each dimension, the resulting empirical clusters from both strands were integrated with the corresponding theoretical tension to formulate one meta-requirement per dimension (five total, Section 3.3). In Phase 3, each meta-requirement was in turn decomposed into two design principles (ten total, Section 4), yielding a traceable path from theory and empirical evidence through design dimensions and meta-requirements to actionable design principles.

## 3. Conceptualising Explanatory AI: theoretical foundations and meta-requirements

This section develops the conceptual foundation for Explanatory AI. We first synthesise multidisciplinary perspectives on human explanation with the evolution of AI explanation capabilities to identify five design dimensions distinguishing Explanatory AI from XAI (Section 3.1). Building on this foundation, we formally define Explanatory AI as an emerging design-paradigm in relation to XAI (Section 3.2). We then examine each design dimension in detail, integrating theoretical tensions with empirical evidence from our healthcare study to derive five corresponding meta-requirements (Section 3.3).

### *3.1. Human explanation principles and evolution of AI explanation*

Explanations constitute the fundamental social and cognitive mechanism through which humans exchange understanding, whether children asking why or scientists justifying theories (Lombrozo 2012; Miller 2019). Across the disciplines we surveyed, three principles recur consistently. First, explanations are causal, locating outcomes within webs of causes, reasons, or mechanisms. Philosophy of science debates whether explanations require logical deduction from general laws (Hempel and Oppenheim 1948) or identification of causal-mechanistic processes (Salmon 1984), with pragmatic approaches emphasising that what counts as explanation depends on questioner context and interests (van Fraassen 1980). Second, explanations are contrastive, answering ‘Why P rather than Q?’ rather than providing comprehensive accounts. Peirce’s (1955) abductive reasoning selects hypotheses making facts unsurprising, while cognitive psychology shows people favour explanations that are simple, coherent with prior beliefs, and highlight abnormal conditions against normal backgrounds (Hilton and Slugoski 1986; Lombrozo 2012). Third, explanations are audience-relative, judged effective based on recipient background and goals. Communication research demonstrates that effective explanations follow principles analogous to Gricean conversational maxims, supplying appropriate detail and avoiding unnecessary jargon (Grice 1975), and writing-clarity guidelines (Williams and Bizup 2017). Education research shows that effective teaching employs analogies, stepwise causal narratives, and comprehension checks, with the self-explanation effect demonstrating how generating explanations fosters learning and reshapes beliefs (Chi et al. 1989; Dahlstrom 2014; Dunlosky et al. 2013; Rosenshine 2012). Social psychology reveals that moral judgment colours explanation, with people perceiving harmful side effects as more intentional than identical helpful ones (Knobe 2003), while folk explanations differ systematically depending on perceived intentionality of behaviour (Malle 2004).

AI explanation approaches have progressed through phases characterised by different technological paradigms and explanation strategies. The foundational expert systems era (1980s–1990s) established rule-based explanation through systems like MYCIN, which demonstrated how production rules enable retrospective explanation of diagnostic recommendations (Shortliffe et al. 1975). Swartout (1983) distinguished explaining programme behaviour from justifying why behaviour is appropriate, arguing effective justification requires access to knowledge and reasoning used during system development. His XPLAIN system and subsequent Explainable Expert Systems framework captured design knowledge for generating strategy explanations and terminology descriptions (Swartout, Paris, and Moore 1991). Clancey (1983) revealed fundamental limitations in treating expert knowledge as uniform associations, establishing that effective explanation requires making explicit deeper knowledge embedded procedurally in rules. The DENDRAL retrospective documented comprehensive approaches to scientific reasoning and explanation in the first expert system (Lindsay et al. 1993).

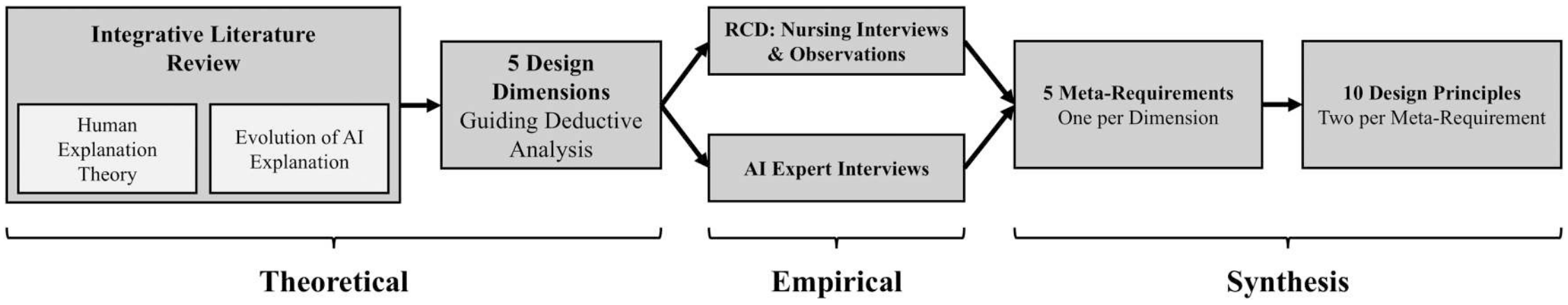

**Figure 1.** Overview of the research approach, showing the flow from theoretical foundations to actionable design guidance.

The statistical and machine learning era (2000s–2010s) brought new challenges as neural networks lacked explicit rules enabling explanation. Craven and Shavlik (1996) pioneered extraction of decision trees from trained networks to approximate reasoning in interpretable form. Štrumbelj and Kononenko (2014) systematically applied Shapley values from game theory to machine learning predictions, providing theoretical foundations for SHAP (Lundberg and Lee 2017), which unified multiple explanation methods. Parallel developments included gradient-based methods like Integrated Gradients addressing saturation problems (Sundararajan, Taly, and Yan 2017) and layer-wise relevance propagation decomposing predictions to individual features (Bach et al. 2015). LIME advanced the field by proposing model-agnostic local explanations through interpretable surrogate models (Ribeiro, Singh, and Guestrin 2016).

The modern generative AI era (2020s-present) has fundamentally transformed explanation capabilities beyond algorithmic transparency towards human-centered communication. Chain-of-thought prompting demonstrated that large language models can generate intermediate reasoning steps improving performance on complex tasks (Wei et al. 2022). Mechanistic interpretability emerged as crucial research direction, with studies reverse-engineering how models implement mathematical and sequential reasoning (Gould et al. 2024; Hanna, Liu, and Variengien 2023). Singh et al. (2023) pioneered hybrid approaches combining LLM capabilities with interpretable models through their Aug-imodels framework, while Singh et al. (2024) argue that LLMs enable redefining interpretability with more ambitious scope through natural language explanation capabilities handling complex patterns beyond traditional methods.

These multidisciplinary principles and technological capabilities together reveal five fundamental dimensions characterising the design space for Explanatory AI systems (see Section 2 and Figure 1 for an overview of how these dimensions were derived from the two literatures), with integrated theoretical and empirical analysis presented in Section 3.3.

### *3.2. Defining explanatory AI: an emerging complementary paradigm*

The evolution traced in Section 3.1 demonstrates a progression from rule-based systems explaining their own reasoning, through statistical methods revealing algorithmic processes, to generative AI systems capable of serving as explanatory partners for human understanding. This progression, combined with multidisciplinary insights showing that effective human explanations are causal, contrastive, and audience-relative, reveals the potential for a fundamentally different approach to AI explanation that we term *Explanatory AI: systems that leverage generative capabilities to serve as explanatory partners for human understanding by providing contextual, narrative-driven explanations that integrate existing domain knowledge, adapt to user characteristics, and prioritise comprehension over algorithmic transparency*. We define Explanatory AI primarily by its explanatory purpose and design objective, supporting human comprehension and informed decision-making, rather than by any particular technology, system class, or interface format. This functional definition includes several more specific readings without being reducible to any one of them: Explanatory AI can manifest as a class of standalone AI systems, as a design paradigm guiding how AI-mediated explanations are constructed, as a user-interface layer translating outputs from other AI systems, as a generative-AI-enabled explanation interface, or as a broader theoretical construct describing AI-mediated sense-making. Throughout this paper, we treat the design paradigm as our primary lens, since our contribution centres on meta-requirements and design principles guiding system construction. The remaining readings describe possible instantiations or theoretical extensions of this paradigm rather than alternative definitions. Unlike traditional AI systems that explain their own decision processes, these systems explain domain knowledge, contextual reasoning, and why recommendations make sense within users' professional contexts.

The claim that Explanatory AI constitutes an emerging paradigm requires clarifying how we use the term. Our use of the term draws on Kuhn's (1970) account of paradigms as organising shared commitments and exemplary problem solutions within a research community, though we follow its subsequent adaptation in HCI, where paradigms need not replace one another. Harrison, Tatar, and Sengers (2007) conceptualise paradigms as broad perspectives that foreground particular phenomena and research questions and shape corresponding approaches to design, evaluation, and knowledge claims. In this HCI-oriented sense, paradigms need not constitute mutually exclusive or successive categories: different perspectives may coexist, overlap, and become relevant to different aspects of the same system. The notion of design paradigms remains salient in contemporary human-AI interaction research, where emerging capabilities such as adaptability, collaboration, bidirectionality, and proactivity motivate reconsideration of established interaction models and design paradigms (Sun et al. 2026).

On this basis, we position Explanatory AI as an emerging design paradigm because it places a distinct explanatory objective at the centre of system design: enabling human understanding and informed action. Understood in this way, XAI is itself a paradigm rather than merely a set of techniques. Whereas model-centered XAI primarily asks how a model arrived at an output and evaluates explanations with respect to the underlying computation, Explanatory AI asks why an output or recommendation makes sense within the user's domain and situational context. The distinction therefore concerns the explanatory purpose that guides design rather than a strict technological boundary. Human-centered, interactive, and conversational XAI already move explanation towards user needs, and Explanatory AI builds on these developments while making human sense-making the organising objective of explanation design. In this respect, the paradigm does not depend on introducing an entirely new set of techniques: as Harrison, Tatar, and Sengers (2007) illustrate, a paradigmatic orientation can also take shape by moving previously marginal concerns to the centre of inquiry and design. XAI and Explanatory AI may consequently operate independently or together within the same system, with different explanatory components serving validation and comprehension purposes respectively. We describe Explanatory AI as emerging to denote a coherent design orientation whose constituent concerns are increasingly visible, but whose systematic articulation around human understanding is still developing.

Explanatory AI complements rather than replaces explainable AI, with each serving different purposes. Both can operate independently or in combination. Some users may require both approaches simultaneously. A radiologist might use XAI saliency maps to validate detected abnormalities (model validation) while using Explanatory AI to understand clinical reasoning and communicate results (comprehension). Similarly, a loan officer might need XAI feature importance for regulatory compliance while using Explanatory AI to explain decisions to applicants. In such combined deployments, XAI maintains algorithmic transparency while Explanatory AI enhances human understanding. Table 1 presents the key distinguishing aspects for Explanatory AI from Explainable AI. Explanatory AI is characterised here primarily as a design paradigm defined by its explanatory purpose; concrete instantiations may take the form of standalone systems, interface layers translating existing AI outputs, or generative-AI-enabled explanation components. The comparison in Table 1 contrasts XAI and Explanatory AI at the level of purpose and design objectives rather than as two fixed, mutually exclusive system classes. The contrasts it draws characterise the prevailing emphasis of each paradigm rather than strict technical boundaries: recent human-centered XAI work already advances interactivity, user adaptation, cognitive-load management, and explanation usefulness (Calisto et al. 2025; Cetinkaya and Krämer 2026; Kaur et al. 2020; Raees et al. 2024; Rong et al. 2024; Schoonderwoerd et al. 2021), and the table therefore contrasts typical XAI implementations with the design emphasis of Explanatory AI, not the outer limits of what XAI can achieve. Explanatory AI does not negate these developments but reorganises and extends them, treating human understanding as the primary design objective and taking contextual, narrative-driven explanation grounded in domain knowledge as its starting point rather than progressively enriching model-tethered explanations.

**Table 1.** Distinguishing explanatory AI from explainable AI.

| Aspect | Explainable AI (XAI) | Explanatory AI |
|---|---|---|
| Primary Purpose | Algorithmic transparency: Model validation, debugging, audit, regulatory compliance | Human understanding: Enabling comprehension, informed decision-making, and knowledge building |
| Primary Question | "How did the model arrive at this output?" | "Why does this make sense?" |
| Typical Users | AI developers debugging models; domain experts validating AI decisions; auditors checking compliance; researchers analysing behaviour | End users requiring understanding; practitioners integrating AI into workflows; professionals making informed decisions; learners building domain expertise |
| Technological Basis | Discriminative AI models and post-hoc XAI methods (e.g. SHAP, LIME) | Typically generative, multimodal models (e.g. LLMs) as enabling technology |
| Output Format | Statistical visualisations, feature importance scores | Multimodal explanations (e.g. text, audio, video) |
| Interactivity | Static presentation of explanations | Interactive presentation and generation of explanations (e.g. dialogue) |
| Epistemic Stance | Prioritises fidelity to algorithmic processes; seeks model-centric accuracy | Prioritises contextual plausibility and domain grounding for comprehension, while remaining bound by fidelity, truthfulness, and domain validity |
| Adaptivity | Uniform outputs in standard implementations; especially suitable for users with statistical literacy | Possible real-time personalisation to user characteristics |

Positioning Explanatory AI within the explanation landscape requires distinguishing it from, rather than opposing it to, several adjacent lines of work that have likewise moved explanation towards the user. *Human-centered XAI* has developed reusable interaction patterns for clinical decision support (Schoonderwoerd et al. 2021) and consolidated a substantial body of user studies on model explanations (Rong et al. 2024), while work on interactive XAI charts the transition from static explanation towards systems that grant users genuine agency (Raees et al. 2024). A parallel strand on *conversational and dialogue-based explanation* treats explanation as a two-way, interactive exchange, from interactive dialogue systems that let practitioners interrogate model predictions in natural language (Slack et al. 2023) to narrative explanation generated with large language models (Martens et al. 2025). Related efforts tailor explanation interfaces to distinct stakeholder groups (Kim et al. 2024) and attend to the information needs of those affected by algorithmic decisions (Schmude et al. 2025), while work on intelligent tutoring systems pursues adaptive, personalised instruction that scaffolds learner understanding, increasingly drawing on AI and language technologies to tailor support to individual needs (Lin, Huang, and Lu 2023). Explanatory AI shares the central commitment animating this work: that explanation must serve human sense-making rather than model transparency alone. It should therefore be read as a shift in emphasis rather than a break from XAI. What distinguishes it is the scope and locus of that commitment. Human-centered and conversational XAI predominantly make the outputs of underlying XAI methods more usable, remaining tethered to a specific model's computations and, in many conversational systems, to a fixed repertoire of preconfigured responses. Explanatory AI instead treats human understanding as the primary design objective and draws on generative and multimodal capabilities to provide contextual, narrative-driven explanation that integrates domain knowledge and need not originate from a particular model's internal processes. These adjacent strands thus form a bridge towards the paradigm we articulate: Explanatory AI reorganises their insights into an explicit set of meta-requirements and design principles and extends them beyond model-tethered explanation, complementing rather than replacing established human-centered XAI along a continuum from model-centered transparency to user-centered sense-making.

Explanatory AI systems can operate in two modes, both prioritising human understanding. *Standalone deployment* serves as explanation partners for domain knowledge independent of specific AI decisions, *while complementary deployment* translates outputs from existing AI decision systems into contextually meaningful explanations. Generative AI capabilities, particularly large language models and transformer architectures (Vaswani et al. 2017), provide the technical foundation enabling dynamic, contextual explanation through narrative-driven content, real-time adaptation, interactive dialogue, and multimodal presentation (Singh et al. 2024; Wei et al. 2022). These capabilities distinguish Explanatory AI from traditional XAI methods, which produce static technical outputs, and general conversational AI systems, which handle open-ended dialogue but lack specialisation in explanation for human understanding (Torkamaan et al. 2024). This conceptualisation establishes Explanatory AI as an emerging design paradigm characterised by five key dimensions that distinguish it from traditional XAI approaches, as detailed in the following section.

A natural question is how to determine, for a given AI system, whether it constitutes XAI, Explanatory AI, or some combination of both. We do not draw this distinction along a single technical/human-centered axis, since generative AI is increasingly used to generate or wrap XAI outputs themselves (e.g. LLM-generated summaries of SHAP values), and a purely technological criterion would therefore misclassify such systems (Martens et al. 2025; Singh et al. 2024). Instead, the distinction is functional: as a paradigm, XAI is operationalised through technical methods (e.g. SHAP, LIME, saliency maps) whose outputs are evaluated against the model's internal computation, regardless of how those outputs are ultimately presented. Explanatory AI, by contrast, operates on top of, or independently of, such methods, as a design layer oriented towards the user's comprehension rather than the model's fidelity. Classifying a concrete system therefore requires asking, for each of the five dimensions in Table 1, which pole its outputs serve: does a given explanation aim to let a developer or auditor verify how the model computed its output (XAI), or does it aim to let a practitioner understand why the output is sensible enough to act on (Explanatory AI)? Many real systems will combine both, as in the loan-officer example above, where the identical underlying decision is paired with a feature-importance report for regulatory audit (satisfying the XAI columns of Table 1: model-centric fidelity, static output, validation purpose) and a narrative explanation for the applicant (satisfying the Explanatory AI columns: contextual plausibility, dynamic communication, comprehension purpose). Because the distinction is one of explanatory purpose rather than system type, it does not partition systems into disjoint classes, and cases that appear to sit between the two are typically dual-purpose deployments in which each explanatory output can still be located by the purpose it serves. For the same reason we do not propose a scale or set of thresholds for assigning a system to one paradigm or the other: the five dimensions in Table 1 function as orienting questions about individual explanatory outputs rather than as measurement criteria for whole systems. Developing validated instruments for characterising where a system's explanatory outputs fall along these dimensions would require dedicated empirical work and remains an open direction for research. The differentiator is thus not the presence of generative technology, but which functional role, validation or comprehension, a given explanatory output is designed to serve.

This distinction also clarifies how Explanatory AI relates to generative AI as a technology. Generative AI (e.g. general-purpose systems such as ChatGPT) is the enabler, not a synonym: producing fluent generated text does not by itself make a system Explanatory AI. A system qualifies only when purposefully designed to satisfy the meta-requirements in Section 3.3, e.g. domain grounding, criticality-calibrated validation, and user adaptation, rather than generating ungrounded, unvalidated output. Generative AI is thus the shared enabler behind both interactive / conversational XAI (e.g. narrating technical outputs) and Explanatory AI (as its primary mechanism), while purpose continues to distinguish the two.

### 3.3. Design dimensions and meta-requirements for explanatory AI

The following subsections present integrated analysis of each dimension characterising Explanatory AI systems, combining theoretical tensions with empirical evidence from our healthcare study (detailed in Section 2: six nursing staff, six AI experts). Each dimension reveals a fundamental design challenge where current XAI approaches fail to meet human explanation needs. Following integrative synthesis methodology (Torraco 2005), we structure each dimension by first presenting the theoretical tension between XAI approaches and human explanation needs, articulating the design challenge arising from this tension, providing empirical evidence from end users and technical experts, and deriving a meta-requirement specifying what Explanatory AI systems must achieve to address the design challenge. These meta-requirements, grounded in multidisciplinary theory and examined against empirical evidence, provide the foundation for our design principles presented in Section 4. The following Figure 2 summarises derived meta-requirements together with their respective design principles, providing structural scaffolding for the following sections.

#### 3.3.1. Explanatory purpose: from diagnostic to interpretive sense-making

This dimension addresses what explanations should accomplish. Traditional XAI methods like SHAP and LIME (Lundberg and Lee 2017; Ribeiro, Singh, and Guestrin 2016) serve diagnostic purposes, revealing which input features contributed to algorithmic decisions for model validation and audit (Adadi and Berrada 2018; Meske et al. 2022). These approaches answer 'How did the algorithm decide?' for stakeholders engaged in model validation, debugging, and regulatory compliance, including AI developers, domain experts validating AI decisions, and auditors. In contrast, Explanatory AI pursues interpretive aims, helping users understand why recommendations make sense within their domain expertise and situational contexts (Miller 2019). This fundamental shift addresses 'Why does this make sense?' for end users who must understand and act on recommendations in professional contexts; and even technically literate users struggle with current XAI outputs (Kaur et al. 2020), suggesting that end users in professional contexts face even greater challenges. The design challenge lies in creating systems that support interpretive sense-making to enable informed decision-making within sociotechnical environments, potentially incorporating pedagogical functions where appropriate (Bahroun et al. 2023; Tlili et al. 2023). This is further complicated by the fact that explanations can simultaneously foster human-AI collaboration and provoke identity threat, depending on how they are designed and the degree to which they signal AI autonomy (Wang, Zhuang, and Santoso 2026).

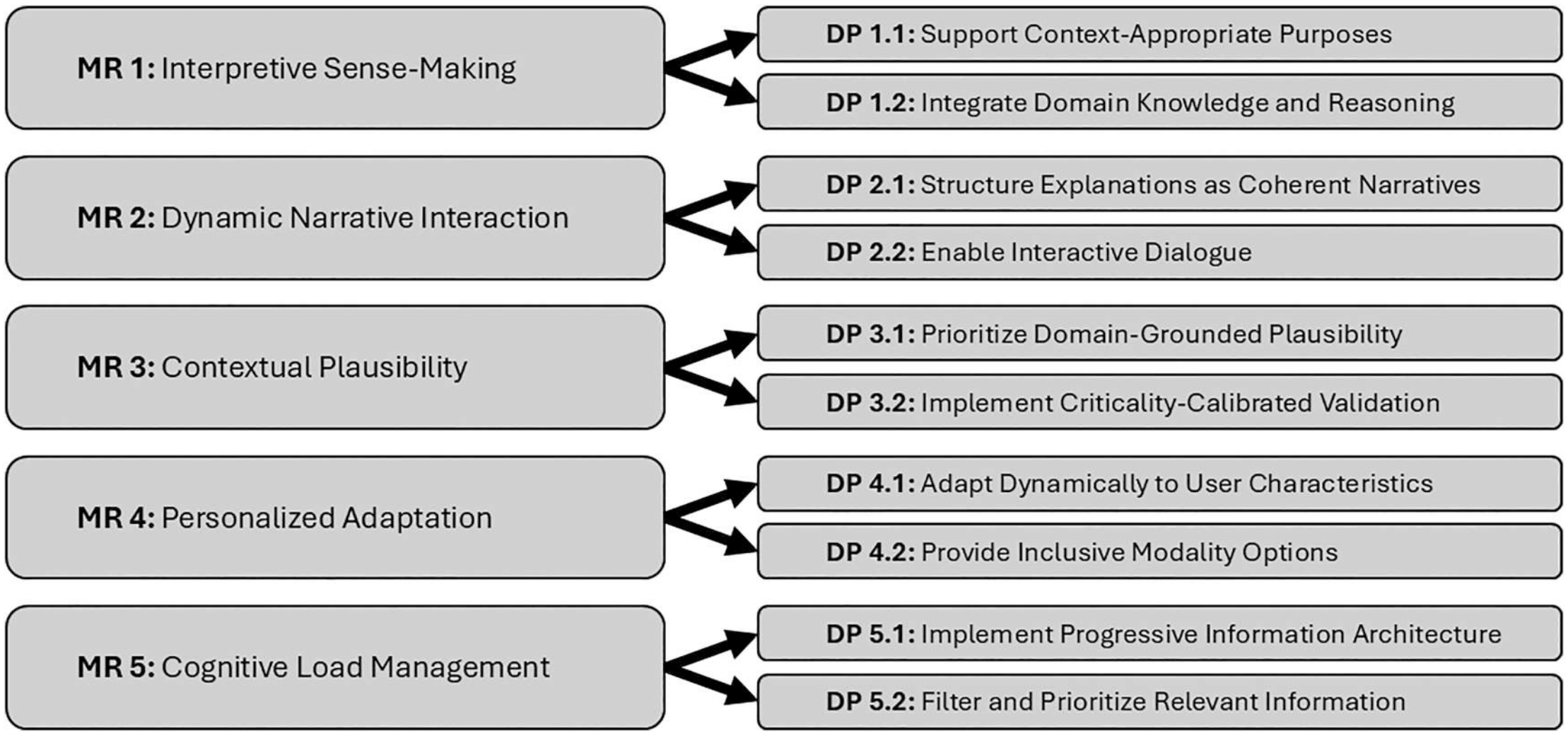

**Figure 2.** Overview of derived meta-requirements (MR) on the left and their respective design principles (DP) on the right side.

Explanatory AI operates in two modes, both prioritising user comprehension. Standalone deployment provides domain knowledge explanations independent of AI decisions. In combination with XAI, Explanatory AI translates technical XAI outputs into contextually meaningful explanations while XAI maintains validation functions. Some expert users may deploy both approaches simultaneously, using XAI for model validation and Explanatory AI for decision integration, with each serving its distinct purpose within a complementary architecture.

Empirical evidence from our healthcare study supported this theoretical distinction. End users consistently emphasised needs beyond technical transparency, seeking explanations supporting practical reasoning and documentation. One nursing staff member noted:

> What's really needed are phrasing aids. I've often seen that untrained staff hesitate to document anything, and when they do, they struggle a lot. Tools that help them explain or justify their documentation would be extremely helpful. (U05-2)

Another reflected: 'Honestly, the digital training wasn't really helpful. You're just left alone afterward anyway' (U02-8). Documentation serves 'as justification or evidence' in audits (U02), with 'explainability helping back up a decision when something happens', especially for less experienced personnel (U01). Users sought scaffolding helping them formulate and justify decisions within specific situations rather than understanding algorithmic feature importance.

AI experts similarly emphasised that explanations should extend beyond algorithmic mechanics. One expert articulated that explanations should help users 'make the decision processes of an AI understandable for humans so that they can comprehend the end result that an AI produces' (AI-Exp-06). Experts noted: 'Because ultimately, these generative AI explanations can also be explained by generative AI itself' (AI-Exp-05). Others defined explainability as enabling 'a specific target person to comprehend and interpret a certain output from AI' (AI-Exp-02), noting it can optimise prompts: 'if you have an explanation instead, then that might help you to clarify or adjust the prompt or understand how such an error could have occurred' (AI-Exp-05).

*3.3.1.1. Design challenge.* Creating systems that shift from answering 'How did the algorithm decide?'

(diagnostic transparency) to addressing 'Why does this make sense?' (interpretive sense-making) while supporting both operational decision-making and pedagogical learning functions.

*3.3.1.2. Meta-requirement 1 (interpretive sense-making).* Explanatory AI systems must prioritise interpretive sense-making that integrates domain knowledge and contextual reasoning to support understanding and informed decision-making. Systems must adapt explanation purpose to user context, providing operational support for immediate decision-making and pedagogical support for learning and skill development.

### 3.3.2. Communication mode: from static technical abstractions to dynamic narrative interaction

This dimension addresses how explanations are communicated. Effective explanation requires narrative communication adapting to audience needs, following Gricean conversational maxims emphasising appropriate detail, relevance, and clarity (Grice 1975), with evidence that accessible metaphors and clear causal narratives improve understanding (Dahlstrom 2014). Classical XAI methods communicate through static, technical outputs including visualisations, feature importance scores, and mathematical relationships (Bach et al. 2015; Lundberg and Lee 2017; Ribeiro, Singh, and Guestrin 2016; Shrikumar, Greenside, and Kundaje 2017). These fixed-format presentations require statistical literacy to interpret and often create cognitive overload by presenting comprehensive information simultaneously regardless of user contexts (Chromik and Schuessler 2020). Human-centered XAI work such as DoReMi starts to translate these technical outputs into multi-modal UI design patterns for clinical explanations, explicitly linking explanation requirements to reusable interaction patterns (Schoonderwoerd et al. 2021), but still assumes fixed explanation components derived from XAI methods rather than dynamic narrative interaction. Research demonstrates that seemingly minor design choices in XAI presentations fundamentally alter user interpretation, with lay users particularly struggling to understand feature importance plots (Okolo et al. 2024; Szymanski, Millecamp, and Verbert 2021). Numerical explanations such as certainty measures can help users identify errors, while verbal explanations foster higher reliance but risk over-reliance, and combining both types amplifies blind trust in AI (Papenkordt, Ngonga Ngomo, and Thommes 2026). Moving beyond static explainability towards interactive AI systems that provide users with genuine agency represents a critical evolution in human-AI interaction design (Raees et al. 2024). Visual formats, while easy to understand, are particularly prone to over-interpretation of information they do not communicate (Xuan et al. 2025). The design challenge involves implementing dynamic, multimodal communication employing narrative reasoning that mirrors human-to-human patterns, incorporating progressive disclosure, interactive dialogue, and contrastive explanations that allow users to control depth according to immediate needs (Miller 2019; Rosenshine 2012).

Empirical evidence illustrated these communication needs. End users emphasised preferences for communication mirroring natural human interaction rather than formal technical presentation, seeking 'adaptable instructions, possibly in plain language' (U01) and 'systems that talk like a human being, not like a manual, to effectively break language barriers' (U06). Formal terminology created psychological distance counterproductive in work environments (U04), with preferences for explanations that 'sound like advice from a colleague' rather than abstract rulebooks (U01, U04). Multimodal formats proved essential: 'What really helped me were videos. It would be great to have access to a system where everything is explained again in that way' (U02-9), and 'Like having some kind of video database, where I can just click and see, this is how you apply a bandage, for example' (U05-5). Observational notes highlighted needs for 'speech to text for documentation' (U05) and visual models 'that you can just follow step by step without having to read long text' (U01).

Experts emphasised that natural language breaks communication barriers: 'Natural language explanations are important for simply speaking to normal consumers, because they can't really comprehend all these processes' (AI-Exp-06), with generative AI offering flexibility: 'Natural language potentially also with images is possible, which is more adaptive compared to XAI outputs, which are more static' (AI-Exp-01). The chat-like format emerged as advantage, with 'interaction in the form of dialogue being simply learned human behavior' (AI-Exp-01). Interactive questioning proved crucial: 'Simply when you apply for a loan, in my opinion it would make sense to say again: yes, why exactly? The more contextual information you have, the more it definitely helps' (AI-Exp-03). However, experts warned about 'GenAI interpreting too much into the data itself, and unfamiliar people might be misled' (AI-Exp-02), with concerns about 'complaisance, meaning that they want to accommodate you, and there is certainly also the risk that embellished answers will be given' (AI-Exp-05).

*3.3.2.1. Design challenge.* Moving from static technical visualisations requiring statistical literacy to dynamic,

multimodal communication that mirrors human-to-human interaction patterns while managing risks of over-interpretation and over-accommodation.

*3.3.2.2. Meta-requirement 2 (dynamic narrative interaction).* Explanatory AI systems must employ dynamic narrative communication using natural language, contrastive framing, multimodal presentation capabilities, and interactive dialogue supporting real-time clarification. Systems must move beyond static technical visualisations towards adaptive communication mirroring human conversational patterns while managing risks of over-interpretation and over-accommodation.

### 3.3.3. Epistemic stance: from strict algorithmic correspondence to contextual plausibility

This dimension addresses explanation accuracy standards. While philosophy of science debates whether explanations require strict logical deduction (Hempel and Oppenheim 1948) or can employ mechanistic accounts (Kitcher 1989; Salmon 1984), evidence from communication and education suggests slightly simplified explanations often prove more effective for learning than technically precise but incomprehensible accounts (Rosenshine 2012). Traditional XAI prioritises fidelity to algorithmic processes, seeking model-centric accuracy in feature attribution, with methods like Integrated Gradients providing completeness guarantees (Sundararajan, Taly, and Yan 2017) and SHAP offering theoretical foundations ensuring accurate feature attribution (Lundberg and Lee 2017). This approach prioritises algorithmic correspondence, ensuring explanations faithfully represent model computations. However, users may benefit from explanations prioritising contextual plausibility over exhaustive algorithmic detail, incorporating domain knowledge and established reasoning patterns while accepting comprehensibility trade-offs where appropriate (Miller 2019). The design challenge involves determining when to prioritise technical precision versus meaningful understanding, recognising that explanatory effectiveness may require surfacing less algorithmic detail, trading completeness rather than fidelity, for enhanced comprehensibility. This flexibility introduces risks of hallucinated or inaccurate explanations (Zhang et al. 2025), and even when explanations are factually accurate, highly comprehensible formats may lead users to infer spurious information beyond the explanation's scope (Xuan et al. 2025). Furthermore, the modality of explanation, numerical vs. verbal, substantially affects whether users achieve appropriate reliance or fall into over-reliance (Papenkordt, Ngonga Ngomo, and Thommes 2026). This requires carefully designed validation mechanisms and contextual adaptation based on application criticality.

Empirical evidence revealed end users prioritising practical comprehensibility over technical fidelity. Technical explanations, even when algorithmically correct, proved unintelligible. Observations emphasised 'concrete examples, not theoretical explanations' (U04) and explanations 'linked to the care context and not to technical principles' (U01, U02). Recurring themes showed preference for instruction that 'makes sense for the current task or patient' rather than abstract model behaviour (U01, U04), with notes that 'technical transparency is irrelevant when users don't understand the concepts behind it anyway' (U01). Users never expressed desires to understand algorithmic mechanics, confirming that formal XAI fidelity addresses wrong comprehension targets.

Experts distinguished between explicit hallucinations and contextual information addition: 'I believe one must distinguish, genAI can say something false, but genAI can also bring in something correct, which wasn't actually part of this original classification' (AI-Exp-02). Hallucinations emerged as primary limitation: 'As a disadvantage we naturally have the hallucinations which could perhaps then also deliver a false explanation. In super-critical applications in medicine, for example, it may be more fatal if one has a false explanation' (AI-Exp-04). One described the fidelity trade-off:

> I think the biggest limitation is actually this: you potentially move away from the raw numbers and we are no longer model-centric, then I would say that the risk just keeps increasing, because you don't know where the information comes from, there's a lack of traceability (AI-Exp-03) noting 'the main risk is that in this context, it's like a scale: the further you go out, the higher the danger scale goes up' (AI-Exp-03). Experts emphasised need for prompting experience to manage these risks (AI-Exp-01).

The expert concerns above point to a distinction that the notion of contextual plausibility alone does not capture: an explanation can be meaningful to a user without being faithful to the model, factually true, or valid within the domain. To make these requirements explicit and avoid conflating them, Table 2 distinguishes five quality criteria that an Explanatory AI explanation may need to satisfy.

These criteria are distinct and must not be conflated. Plausibility concerns whether an explanation is meaningful to the user and thereby supports comprehension and uptake, but it is one factor among several rather than a precondition: explanations may be comprehensible yet initially implausible, particularly where they challenge prior beliefs. Nor does plausibility substitute for evidential grounding, since an explanation that is plausible yet unfaithful, untrue, or domain-invalid invites overtrust precisely where stakes are highest. Fidelity, unlike the other criteria, presupposes an explicit reference object, which differs by deployment mode: in complementary deployment it is the decision logic of the underlying model, so that factors presented as decisive were in fact decisive for its output; in standalone deployment it is the source evidence and domain knowledge base on which the explanation draws, so that its claims remain traceable to those sources. It does not refer to the internal computation of the generative model producing the explanation text, nor to the internal coherence of the reasoning presented to the user, which is a matter of plausibility rather than correspondence. This reference object also fixes the boundary between acceptable simplification and unfaithfulness. Omitting detail, such as computational steps or secondary factors, reduces completeness rather than fidelity as long as what remains holds with respect to the reference object; restating that object in coarser form is likewise acceptable where the decision-relevant structure is preserved and the approximation is signalled. Asserting reasons or causal relations the reference object does not support is misrepresentation and is never acceptable. The test is therefore not how much is said but whether what is said holds. Fidelity, truthfulness, and domain validity accordingly remain non-negotiable as criticality rises; what narrows is the admissible degree of reduction, and which factors must always be surfaced is a determination for domain experts and the deploying organisation rather than the system itself.

**Table 2.** Distinguishing quality criteria for explanations in explanatory AI.

| Criterion | What it establishes | Reference point |
|---|---|---|
| Plausibility | Whether the explanation is meaningful and makes sense to the user | User comprehension and context |
| Fidelity | Whether the explanation corresponds to the evidential basis it claims to represent | Deployment-dependent: the underlying decision model (complementary) or the source evidence and domain knowledge base (standalone) |
| Truthfulness | Whether the explanation is factually correct | External facts |
| Domain validity | Whether the explanation aligns with accepted professional knowledge | Domain / clinical standards |
| Actionability | Whether the explanation supports responsible, informed decision-making | The decision task |

What varies with application criticality is not whether these criteria apply, but how much reduction of detail is admissible and how strictly compliance is verified. In low-stakes or educational contexts, explanations may remain at a coarse level of detail, and proportionate verification against the reference object is sufficient. As criticality rises, admissible reduction narrows and verification tightens: in high-stakes clinical decisions, an explanation must not only be meaningful but demonstrably traceable to its reference object, factually correct, and consistent with clinical evidence, with validation mechanisms scaled accordingly (see Section 4.3, Design Principle 3.2).

*3.3.3.1. Design challenge.* Balancing technical precision with meaningful understanding by determining when to prioritise contextual plausibility over exhaustive algorithmic detail while managing hallucination risks through validation mechanisms calibrated to application criticality.

*3.3.3.2. Meta-requirement 3 (contextual plausibility).* Explanatory AI systems must prioritise contextual plausibility over exhaustive algorithmic detail where this serves comprehension, while treating fidelity to the applicable reference object, truthfulness, and domain validity as independent requirements that plausibility does not replace. Systems must implement careful validation mechanisms and contextual adaptation based on application criticality, with high-stakes domains requiring heightened safeguards against hallucination risks.

### 3.3.4. Adaptivity: from uniform expert-oriented design to personalised universal accessibility

This dimension addresses explanation personalisation across user populations. Evidence that individual differences in prior knowledge shape causal attribution and explanation selection (Hilton and Slugoski 1986; Knobe 2003), combined with Gricean principles emphasising appropriate audience adaptation (Grice 1975), demonstrates that effective explanations must adapt to recipient characteristics. Classical XAI systems generate uniform outputs in standard implementations, designed primarily for users possessing statistical literacy, with limited personalisation capabilities in typical deployments (Meske et al. 2022). Research demonstrates dramatic performance differences between expert and lay users interpreting identical explanation formats, with lay users achieving significantly lower accuracy (Szymanski, Millecamp, and Verbert 2021), and even within a single professional group, different expertise levels respond differently to AI communication styles, benefiting from varying degrees of assertive versus suggestive guidance (Calisto et al. 2025). Similarly, in healthcare, clinicians and patients interacting with the same underlying system require distinct explanation interfaces (Kim et al. 2024). Information needs of AI novices differ systematically from those of domain experts (Schmude et al. 2025). The one-size-fits-all approach systematically excludes diverse user populations and limits accessibility. Conjoint analysis of user preferences further reveals that users are not uniform in their attitudes towards XAI: some prioritise high explainability, others prioritise fairness and reliability over explainability, indicating that any single design strategy will fail a substantial user segment (Cetinkaya and Krämer 2026). Beyond functional adaptation, different explanation types, such as counterfactual versus feature importance explanations, affect users' psychological distance from the AI (Wang, Zhuang, and Santoso 2026). The design challenge involves implementing personalisation that adjusts vocabulary complexity, conceptual depth, and cultural references to diverse user populations (Calisto et al. 2025; Schmude et al. 2025) while ensuring inclusive accessibility, including accommodations for users with disabilities (Trewin et al. 2019). Recent work on personalised AI communication in clinical decision support further shows that adaptation must also include communication tone and degree of assertiveness, as interns and junior clinicians benefit from more directive guidance while senior clinicians prefer collaborative, suggestive communication styles (Calisto et al. 2025).

Empirical evidence demonstrated adaptation needs. End users emphasised systems must adapt to individual skill levels, preferences, and routines: 'It would be helpful if the system could adapt to the skill level of the caregiver' (U01-6) and 'Older caregivers often have trouble with technology. It would make a big difference if the system could adjust to their level of experience' (U02-10). Observations reinforced adaptation needs: 'each caregiver has a different approach to documentation and information retrieval' (U03), with some 'more visually oriented' while others 'rely on written text for verification' (U04). Daily routines affected engagement, with 'more detailed explanations tolerated during quieter evening shifts' whereas 'during busy times, users only want the bottom line' (U01). Linguistic and cultural diversity emerged as critical: 'At the moment, we have four nursing trainees from India. Despite ongoing German lessons, noticeable language barriers persist' (U06-4), with strong translation requests (U06-6). Notes highlighted how 'newly immigrated staff often hesitate to ask questions due to language insecurity' (U01, U04, U05, U06), and 'some staff avoid documentation tasks entirely due to fear of making linguistic errors' (U04).

Experts noted XAI's limited personalisation: 'In my opinion, the personalization of XAI outputs is virtually non-existent. I've tried out various things, Gradcam, SHAP, LIME, and the customization options are minimal. Maybe I can change a colour in a plot' (AI-Exp-03). Generative AI's flexibility emerged clearly: 'Different stakeholders have different needs for explanations' (AI-Exp-02), with 'simply through this flexibility, one can individually adapt the outputs of generative AI to the person it is addressed to' (AI-Exp-04). However, personalisation requires establishment: 'you always have the problem that the AI doesn't really know what my knowledge level is' (AI-Exp-06). Experts highlighted inclusivity benefits: 'I believe that generative AI definitely has the greater advantages, especially regarding plots, which one perhaps also cannot read aloud, that means someone who is blind or has a visual impairment' (AI-Exp-04).

*3.3.4.1. Design challenge.* Moving from one-size-fits-all uniform outputs to real-time personalisation that adapts to user expertise, language proficiency, cultural backgrounds, and disabilities while ensuring universal accessibility across diverse populations.

*3.3.4.2. Meta-requirement 4 (personalised adaptation).* Explanatory AI systems must provide real-time adaptation to user expertise, language proficiency, cultural backgrounds, and disabilities. Systems must personalise communication by adjusting vocabulary, complexity, and modality while ensuring inclusive accessibility across diverse populations, moving beyond one-size-fits-all approaches towards universally accessible explanation.

### 3.3.5. Cognitive design: from information overload to cognitively-aligned delivery

This dimension addresses information structuring for comprehension. Cognitive psychology demonstrates working memory limitations constrain information processing (Sweller 1988), while education research identifies specific techniques that optimise learning outcomes (Dunlosky et al. 2013). Communication theory emphasises providing just enough detail while avoiding overwhelming recipients (Grice 1975). Classical XAI approaches prioritise complete representation of algorithmic decision processes through comprehensive feature attributions presented simultaneously. When these technical visualisations are presented to users lacking statistical literacy or appropriate contextual framing, they often create cognitive overload rather than enhancing understanding (Bove et al. 2024; Chromik and Schuessler 2020; Schmude et al. 2025). This creates a transparency-comprehension paradox where comprehensive information exposure actually decreases rather than enhances comprehensibility. Research implementing XAI following HCI principles demonstrates careful design can reduce cognitive load (Mohseni, Zarei, and Ragan 2021; Rong et al. 2024), yet traditional approaches often overwhelm users through simultaneous presentation of extensive feature attributions, confidence intervals, and uncertainty measures. The design challenge involves structuring explanations to minimise cognitive load while optimising comprehension, recognising that effective explanation prioritises clarity over information exposure and that excessive transparency can harm trust and decision-making quality (Bove et al. 2024; Miller 2019).

Empirical evidence revealed significant cognitive overload challenges. End users described autopilot behaviour under exhaustion: 'If I work ten shifts in a row, I just go on autopilot' (U04-3), with generational barriers: 'I think it's harder for the older generation. They often struggle with using technology' (U02-10). Observations captured how information in long text blocks overwhelms users under time pressure (U02), emphasising 'keeping steps short and clear' (U03) where fragmented attention leads to errors if interfaces are too dense. Several notes mentioned systems being cluttered and unintuitive (U03, U04, U06), indicating linear, text-heavy designs prove counterproductive. Cumulative data revealed tensions: 'users often skip long texts and only read the first sentence' (U01), with 'too much explanation leading to more confusion, not less' (U04) particularly under pressure. Related entries noted 'users prefer being guided than being shown everything at once' (U01, U05), with some feeling more confident after receiving 'short and confident explanation', even without fully understanding system logic (U04).

Experts articulated overload effects explicitly: 'If I just get 30 plots slapped down from XAI and first have to search through them myself, then at some point I maybe don't want to deal with it anymore, then I'm overwhelmed' (AI-Exp-02). Even with experience, XAI explanations prove 'rather overwhelming and with no experience rather incomprehensible. If one has context and somewhat more knowledge about the topic, then it could make sense' (AI-Exp-01). Generative AI can help filter overload: 'I believe generative AI can definitely help to filter this out, such as a summary and additionally then also individually helping, if I have additional questions, to interact with it in writing' (AI-Exp-02). Progressive disclosure emerged as solution: 'I think it would definitely help if the information were presented progressively. I also think it would help to make it interactive, because with XAI output, you're usually overwhelmed at first. In my opinion, it's complete information overload' (AI-Exp-03). One described the trust paradox: receiving '30 plots slapped down' leads to being 'overwhelmed and angry at the AI, and then I don't trust it either way' (AI-Exp-02).

*3.3.5.1. Design challenge.* Resolving the transparency-comprehension paradox by structuring information to minimise cognitive load while optimising comprehension, prioritising optimal clarity over comprehensive exposure while recognising that excessive transparency can decrease understanding and trust.

*3.3.5.2. Meta-requirement 5 (cognitive load management).* Explanatory AI systems must employ layered information structure with progressive disclosure, relevance-based information filtering, and cognitive load management. Systems must resolve the transparency-comprehension paradox by prioritising clarity over maximum information exposure, recognising comprehensive information presentation can decrease understanding and trust.

## 4. Design principles for building explanatory AI systems

Section 3 established Explanatory AI as an emerging design paradigm distinguishing itself from traditional XAI through five key dimensions and derived corresponding meta-requirements specifying what systems must achieve. This section translates these meta-requirements into actionable design principles prescribing how to build Explanatory AI systems. While XAI methods serve algorithmic transparency for validation purposes, these Explanatory AI principles address end-user needs: understanding why recommendations make sense within domain contexts. The ten design principles, formulated following design science research methodology (Gregor, Chandra Kruse, and Seidel 2020), provide prescriptive guidance across five fundamental aspects: interpretive sense-making, dynamic narrative interaction, contextual plausibility, personalised adaptation, and cognitive load management. These five aspects are best understood as interdependent layers of a single explanatory architecture rather than mutually exclusive principles. Personalised adaptation (MR4), for instance, necessarily shapes how narratives are structured (MR2) and how information is progressively disclosed (MR5). A system adapting to a novice user's expertise level will, by design, also alter narrative pacing and information layering. We separate them analytically to derive distinct, actionable design guidance for each concern, but practical implementations will typically address multiple layers simultaneously, and design principles from different meta-requirements are expected to jointly inform individual interface elements.

In the following subsections, we introduce pairs of design principles derived from each meta-requirement, accompanied by Figma mockups that illustrate one possible implementation in mobile Explanatory AI interfaces supporting nursing care tasks. These mockups are illustrative design artefacts created by the authors to visualise how the design principles could be operationalised; they have not been subjected to user evaluation or expert validation, and we do not claim they demonstrate the principles' effectiveness. Mockups were created by the authors using Figma, grounded in the healthcare scenarios encountered during our contextual interviews. For each design principle pair, we first identified a representative care task from our field data (e.g. wound documentation, medication safety checks, shift handover), then iteratively designed interface elements explicitly instantiating the corresponding design principles, checking each element against the principle's stated implementation guidance to ensure traceable correspondence between principle and design feature. This internal correspondence check was conducted by the author team and does not constitute user or expert validation. While we present the principles in pairs for clarity of derivation, they are intended to operate together within a single Explanatory AI system rather than in isolation; their interactions and the trade-offs they entail are discussed in Section 5.1. Each principle below is stated in a consistent format specifying the user context (C), the mechanism (M) the system should provide, and the goal (G) to be achieved. The theoretical and empirical rationale (R) follows in a dedicated paragraph labelled *Rationale*, complemented by implementation guidance.

Table 3 provides an overview of the derivation chain, linking a representative empirical theme to its underlying theoretical construct, the associated design challenge, the corresponding meta-requirement, and the resulting design principle for all ten principles introduced below.

## 4.1. Design principles for interpretive sense-making (MR1)

### 4.1.1. Design principle 1.1 (support context-appropriate purposes)

For care workers who need immediate decision support during time-pressured tasks or skill development during quieter periods, the system should provide purpose-switchable explanation modes (operational vs. pedagogical), detected or user-specified, to achieve effective interpretive sense-making adapted to the moment's demand.

*4.1.1.1. Rationale.* Meta-Requirement 1 established that systems must support interpretive sense-making in professional contexts (Section 3.3.1). Empirical evidence showed users need different types of explanations depending on context: scaffolding 'to help them explain or justify their documentation' (U05-2) for operational tasks, versus learning support as 'the digital training wasn't really helpful' (U02-8) indicates pedagogical needs. Regarding potential didactical use AI experts noted, that 'if you have an explanation instead, then that might help you to clarify or adjust the prompt or understand how such an error could have occurred' (AI-Exp-05). Users sought both immediate task support and longer-term understanding.

*4.1.1.2. Implementation.* Design systems that detect or allow users to specify whether they need operational support (making/justifying immediate decisions) or pedagogical support (learning domain knowledge and building skills). For operational contexts, prioritise actionable guidance and justification support enabling decision documentation. For pedagogical contexts, provide progressive learning adjusting complexity over time, worked examples, and explanations building transferable understanding. Enable seamless switching between purposes as user needs change.

### 4.1.2. Design principle 1.2 (integrate domain knowledge and reasoning)

For practitioners evaluating AI recommendations against their professional judgment, the system should provide explanation content grounded in domain ontologies, established practices, and professional reasoning patterns rather than algorithmic mechanics, to achieve recommendations that make sense within users' existing conceptual frameworks.

*4.1.2.1. Rationale.* Meta-Requirement 1 requires integrating domain knowledge to support understanding within professional contexts. Users sought explanations 'linked to the care context and not to technical principles' (U01, U02), with experts emphasising explanations must help users 'comprehend the end result' (AI-Exp-06) through domain-appropriate reasoning rather than algorithmic mechanics.

**Table 3.** Overview of relation between empirical themes, theoretical constructs, design challenges, meta requirements, and design principles.

| Empirical Theme (anchor quote) | Theoretical Construct | Design Challenge | Meta-Requirement/Design Principle |
|---|---|---|---|
| 'to help them explain or justify their documentation' (U05-2) | Explanation purpose | 'How did it decide' vs. 'Why does this make sense' | MR1: DP 1.1 Support Context-Appropriate Purposes |
| 'linked to the care context and not to technical principles' (U01, U02) | Domain-grounded sense-making | | MR1: DP 1.2 Integrate Domain Knowledge and Reasoning |
| 'talk like a human being, not like a manual' (U06) | Gricean maxims, narrative coherence | Static technical output vs. dynamic narrative | MR2: DP 2.1 Structure Explanations as Coherent Narratives |
| 'interaction in the form of dialogue being simply learned human behavior' (AI-Exp-01) | Interactive dialogue, real-time clarification | | MR2: DP 2.2 Enable Interactive Dialogue |
| 'concrete examples, not theoretical explanations' (U04) | Algorithmic correspondence vs. completeness of detail | Precision vs. comprehensibility trade-off | MR3: DP 3.1 Prioritise Domain-Grounded Plausibility |
| 'the further you go out, the higher the danger scale goes up' (AI-Exp-03) | Hallucination/Bias risk, criticality-calibrated validation | | MR3: DP 3.2 Implement Criticality-Calibrated Validation |
| 'adapt to the skill level of the caregiver' (U01-6) | Audience adaptation, individual differences | 'One-size-fits-all' vs. personalisation | MR4: DP 4.1 Adapt Dynamically to User Characteristics |
| 'who are blind or have visual impairment' (AI-Exp-04) | Universal accessibility | | MR4: DP 4.2 Provide Inclusive Modality Options |
| 'often skip long texts and only read the first sentence' (U01) | Working memory limits, progressive disclosure | Transparency vs. comprehension | MR5: DP 5.1 Implement Progressive Information Architecture |
| 'too much explanation leading to more confusion' (U04) | Relevance filtering | | MR5: DP 5.2 Filter and Prioritise Relevant Information |

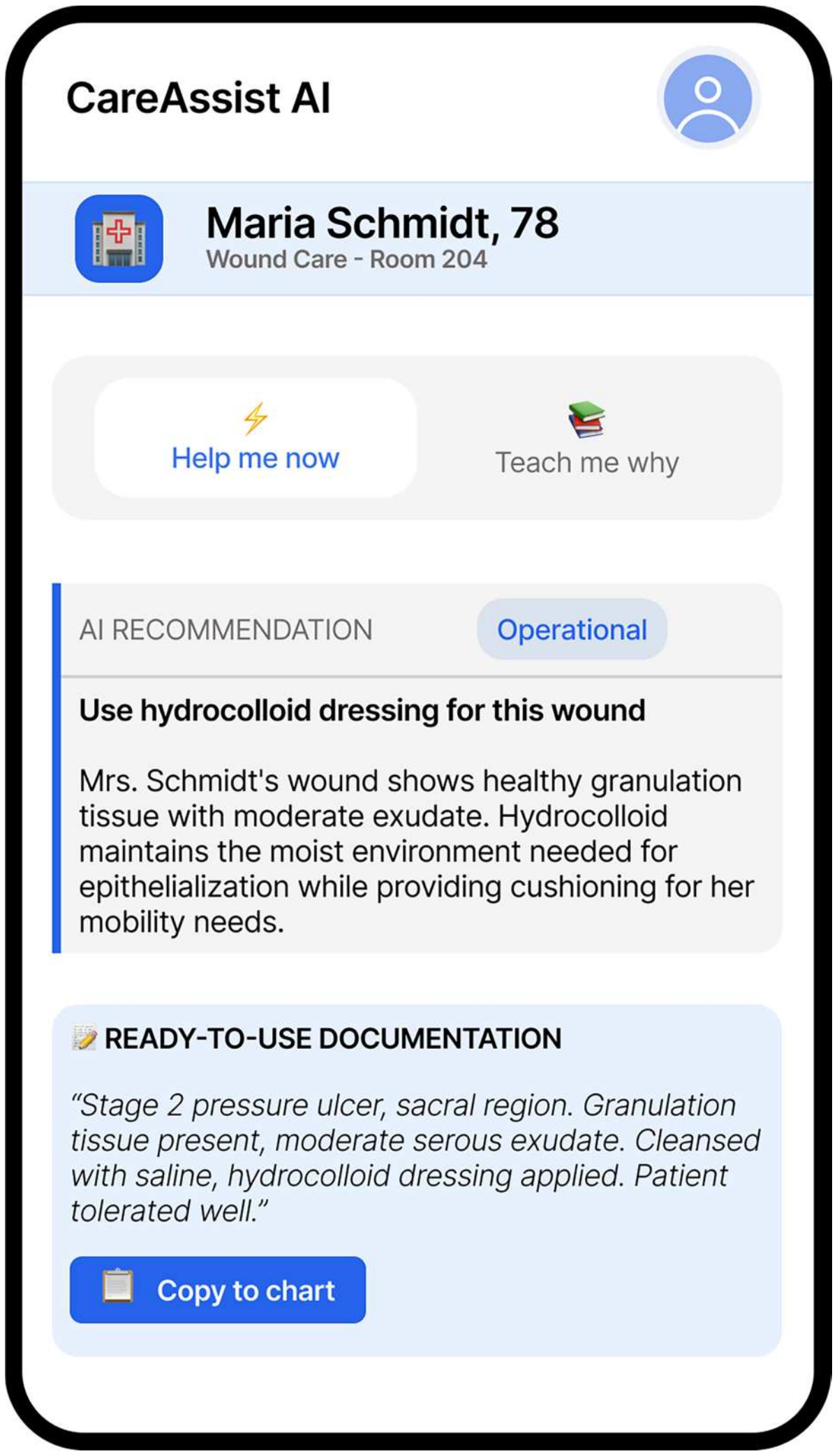

**Figure 3.** Mockup illustrating design principles 1.1 & 1.2 (CareAssist AI – wound care).

*4.1.2.2. Implementation.* Incorporate domain ontologies, established practices, and professional reasoning patterns into explanation generation. Use domain-appropriate terminology and causal language. Frame recommendations within familiar conceptual frameworks. Support documentation and justification needs by helping users articulate their reasoning. Consider worked examples and analogies building transferable understanding.

Figure 3 presents a mockup of CareAssist AI, a wound care assistant that demonstrates the design principles for interpretive sense-making (MR1). The interface displays a patient's wound care information with two interaction modes: 'Help me now' for immediate operational support and 'Teach me why' for pedagogical learning. The system implements DP 1.1 (Support Context-Appropriate Purposes) through these dual modes, allowing users to choose between action-oriented and learning-oriented assistance. DP 1.2 (Integrate Domain Knowledge and Reasoning) is demonstrated through the AI recommendation for hydrocolloid dressing, which includes clinical reasoning about the wound's granulation tissue and exudate levels, along with ready-to-use documentation text.

### 4.2. Design principles for dynamic narrative interaction (MR2)

#### 4.2.1. Design principle 2.1 (structure explanations as coherent narratives)

For users evaluating explanations in the course of everyday professional communication, the system should provide narrative-structured output, causal chains, contrastive framing, domain-appropriate storytelling, to achieve explanations that mirror natural human-to-human explanatory communication.

*4.2.1.1. Rationale.* Meta-Requirement 2 established need for dynamic narrative communication (Section 3.3.2). Users emphasised preferences for systems that 'talk like a human being, not like a manual' (U06) and explanations that 'sound like advice from a colleague' (U01, U04), with experts noting that 'natural language explanations are important for speaking to normal consumers' (AI-Exp-06).

*4.2.1.2. Implementation.* Employ natural language generation producing conversational explanations without unnecessary technical jargon. Structure explanations as coherent narratives with clear causal chains connecting inputs to outputs. Use analogies to familiar concepts. Provide contrastive explanations highlighting 'why this rather than that'. Adapt tone and style to professional context while maintaining accessibility.

#### 4.2.2. Design principle 2.2 (enable interactive dialogue)

For users who need to resolve uncertainty or probe an explanation's reasoning, the system should provide conversational follow-up and clarification capability with validation safeguards against hallucination/over-accommodation, to achieve real-time, dialogic sense-making without introducing new risk.

*4.2.2.1. Rationale.* Meta-Requirement 2 requires interactive dialogue supporting real-time clarification. Experts highlighted 'interaction in the form of dialogue being simply learned human behavior' (AI-Exp-01), with users needing ability to ask 'yes, why exactly?' to gain contextual understanding (AI-Exp-03). However, systems must manage risks of 'GenAI interpreting too much' or 'embellished answers' (AI-Exp-02, AI-Exp-05). Nurses preferred explanations, which 'sound like advice from a colleague', rather than having abstract rulebooks (U01, U04).

*4.2.2.2. Implementation.* Implement conversational capabilities supporting follow-up questions and clarification requests. Maintain conversation histories enabling reference to previous explanations. Allow users to drill down into specific aspects interactively. Design safeguards against over-accommodation and hallucination through validation, particularly in high-stakes domains. Enable users to control dialogue depth according to immediate needs.

Figure 4 shows two panels of DocuCare AI, a conversational documentation assistant for care workers performing evening assessments. The first panel illustrates DP 2.1 (Structure Explanations as Coherent Narratives) through a progressive workflow that guides users through Assessment, Actions, and Plan stages in a logical sequence. The AI initiates the documentation process with 'Let's document Mr. Weber's evening assessment. How was his condition?', creating a natural dialogue flow. DP 2.2 (Enable Interactive Dialogue) is evident in the conversational interface where the nurse provides input ('He seemed more confused today, kept asking about his wife. BP a bit high'.) and the AI responds with structured draft notes while offering clarifying questions ('What was the exact BP?' and 'Other symptoms?') to ensure comprehensive interaction. The second panel illustrates how this dialogue continues dynamically across turns: in response to the AI's clarifying question, the nurse supplies the missing detail ('Patient had no other symptoms, BP was 127/78. Finalise the draft for documentation after updating'.), and the AI incorporates this information into an updated, finalised note rather than treating the initial draft as fixed. This progression demonstrates the adaptive, dialogic sense-making central to DP 2.2: the system maintains conversational state, responds to follow-up input, and revises its output accordingly, allowing the nurse to resolve uncertainty through interaction before finalising documentation.

### 4.3. Design principles for contextual plausibility (MR3)

#### 4.3.1. Design principle 3.1 (prioritise domain-grounded plausibility)

For users who need explanations to guide action rather than verify computation, the system should provide domain-grounded causal reasoning that need not reproduce the full algorithmic detail of an underlying model but must remain faithful to its reference object (Section 3.3.3), to achieve contextual plausibility that supports real-world decision-making and accepts reduced completeness rather than reduced correctness.

*4.3.1.1. Rationale.* Meta-Requirement 3 established need to balance completeness of algorithmic detail with comprehensibility (Section 3.3.3). Users prioritised practical comprehensibility, seeking 'concrete examples, not theoretical explanations' (U04) with explanations 'linked to the care context' (U01, U02). Technical transparency proved 'irrelevant when users don't understand the concepts behind it anyway' (U01), with users never expressing desires to understand algorithmic mechanics. AI experts stated, that 'you potentially move away from the raw numbers and are no longer model-centric, the risk just keeps increasing, because you don't know where the information comes from, there's a lack of traceability', highlighting that reduced detail must not come at the cost of traceability: the continuum concerns how much is surfaced, not whether what is surfaced holds.

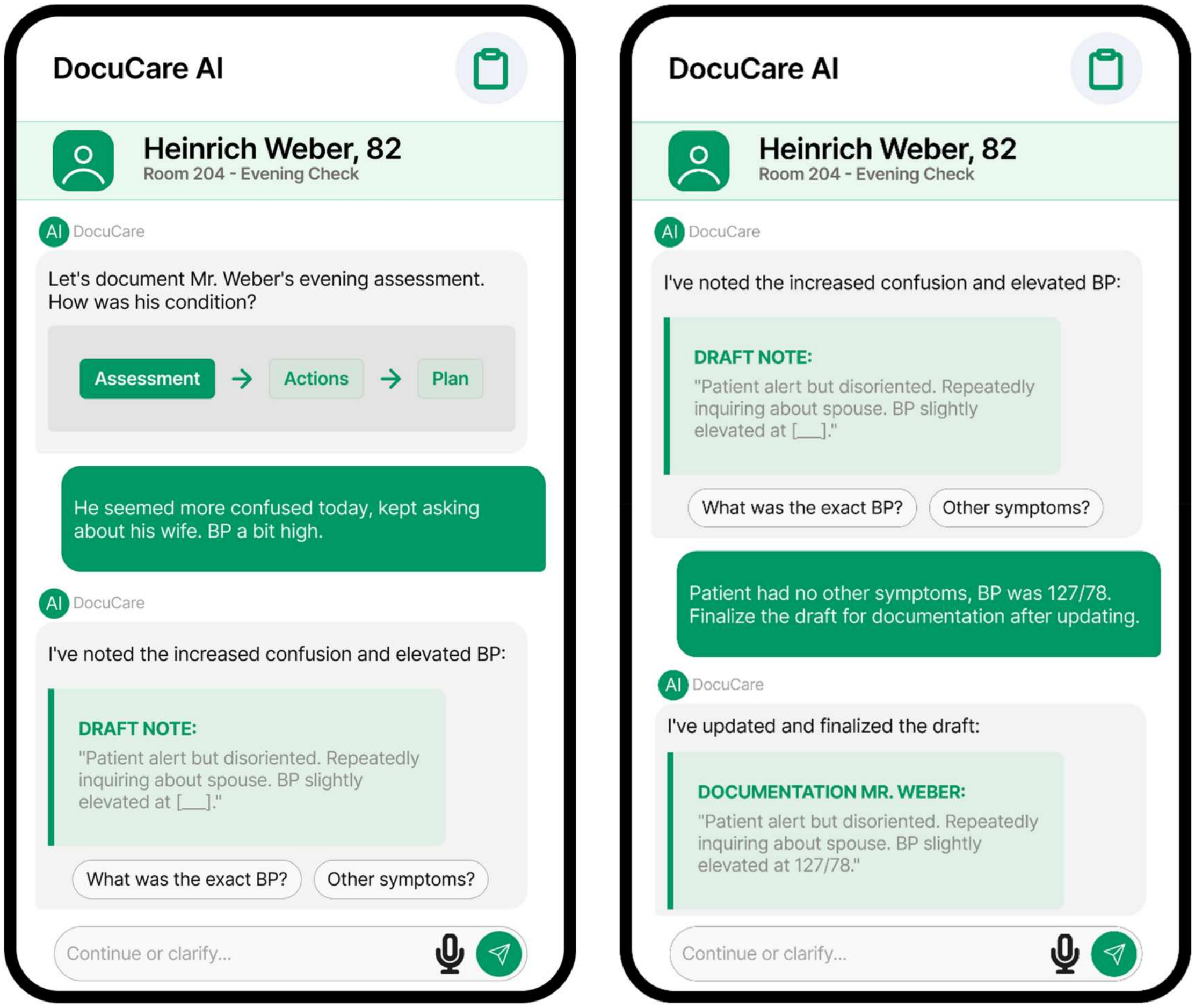

**Figure 4.** Mockup illustrating design principles 2.1 & 2.2 across two sequential dialogue turns (DocuCare AI – documentation).

*4.3.1.2. Implementation.* Incorporate domain knowledge and established reasoning patterns beyond pure algorithmic outputs. Use domain-appropriate causal language and reasoning structures users can evaluate against their expertise. Provide meaningful reasoning rather than exclusively statistical attributions. Accept that effective explanation may prioritise contextual sense-making over completeness of algorithmic detail, while remaining faithful to its reference object. Simplification that remains faithful, factually true, and domain-valid may reduce what is surfaced but must never misstate it: omitting a secondary factor is acceptable, asserting a reason the reference object does not support is not.

#### 4.3.2. Design principle 3.2 (implement criticality-calibrated validation)

For users operating in domains of varying criticality (e.g. medication safety vs. routine documentation), the system should provide validation rigour calibrated to stakes, hallucination detection, consistency checks, hybrid XAI/generative architectures for high-stakes cases, and withholding or deferral to human confirmation where an explanation cannot be verified, to achieve validation rigour and admissible reduction proportional to the risk of harm.

*4.3.2.1. Rationale.* Meta-Requirement 3 requires managing risks through appropriate validation (Section 3.3.3). Experts warned that 'the further you go out, the higher the danger scale goes up' (AI-Exp-03), with hallucinations being 'more fatal in super-critical applications in medicine' (AI-Exp-04). Risk management must scale with application stakes. Nurses furthermore highlighted the need for explanations 'linked to the care context' (U01, U02) and ones, which 'make sense for the current task or patient' (U01, U04).

*4.3.2.2. Implementation.* Assess application criticality and implement appropriate validation rigour. Criticality is a property of the deployment context rather than of the system, so its classification is a governance decision of the deploying organisation together with domain stakeholders, ideally adopting risk categories the domain already operates rather than introducing a parallel scheme. The designer's responsibility is to make those levels operative: to expose criticality as an explicit configuration, bind validation behaviour to it, and document which factors must always be surfaced at each level. Rather than fixed acceptance thresholds, which cannot be specified generically, the principle prescribes the direction of scaling: as criticality rises, admissible simplification narrows, tolerance for unsourced content falls, provenance and uncertainty must be surfaced more explicitly, and the requirement for human confirmation before action tightens.

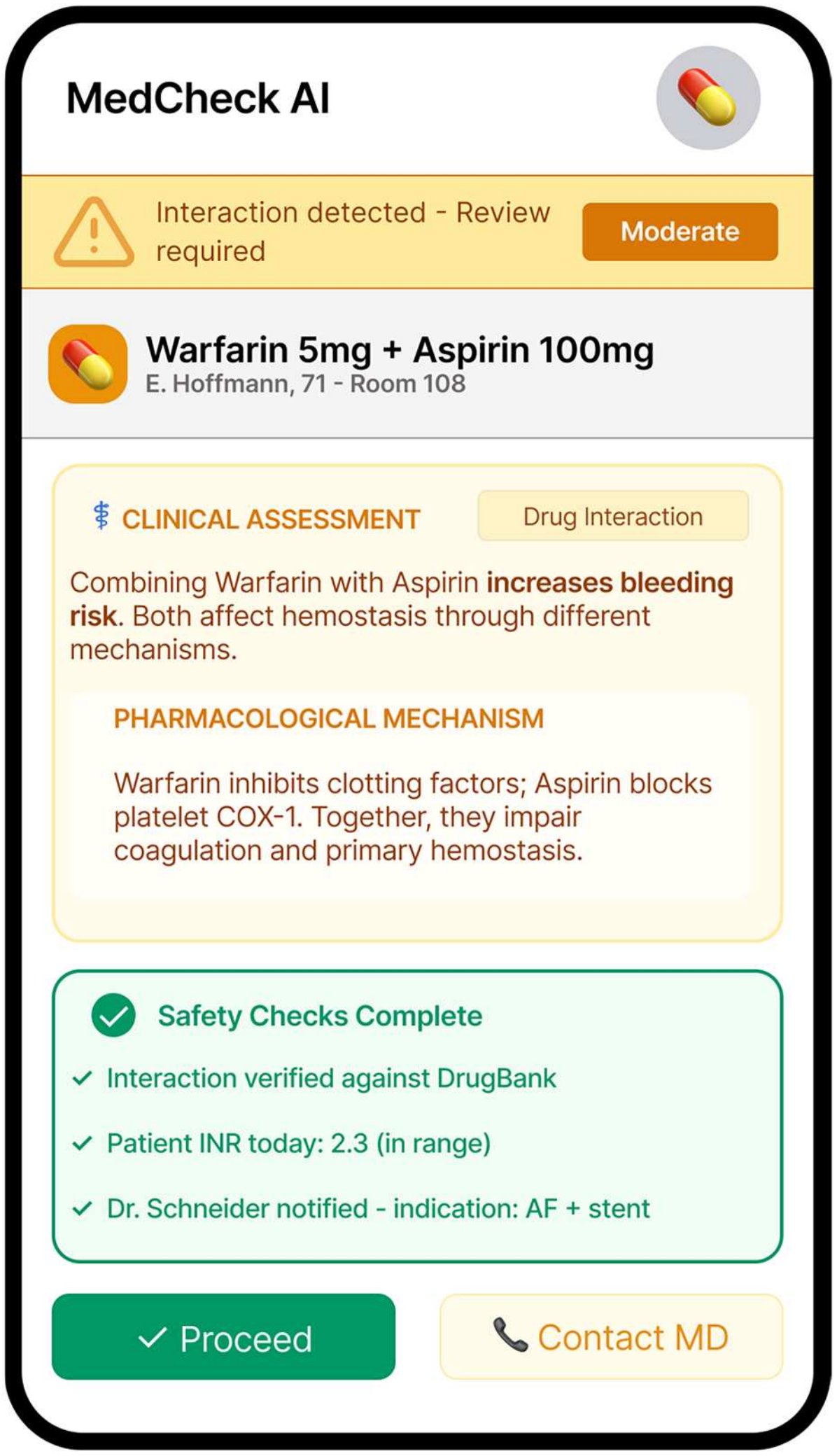

**Figure 5.** Mockup illustrating design principles 3.1 & 3.2 (MedCheck AI – medication interaction).

Consider hybrid architectures combining XAI components for audit transparency with generative components for user comprehension. Provide confidence indicators when explanations extend beyond direct model outputs. Validation operates on two complementary levels. Technically, systems can cross-check generated explanations against model outputs and structured domain sources, apply factual-consistency and hallucination detection, surface uncertainty and provenance, and withhold or flag content that extends beyond verifiable grounds. These safeguards are necessary but not sufficient, because comprehensible explanations can induce automation bias and overtrust regardless of their accuracy. Systems must therefore also support critical engagement on the human level: signalling explanation confidence and limitations explicitly, prompting users to verify high-stakes claims against authoritative sources, preserving access to the underlying evidence, and supporting the AI literacy and awareness that enable users to question rather than defer to plausible output.

Where an explanation cannot be verified against its reference object, the default is to withhold rather than present: unverifiable content is suppressed, marked as unverified, or replaced by the sourced material it was meant to elaborate, with deferral to human confirmation in high-criticality settings rather than silent degradation. Responsibilities extend beyond runtime. Before deployment, domain experts define the knowledge base and which content may be generated at all, reducing hallucination risk by restriction rather than downstream detection alone; during operation, withheld and flagged cases are logged and made reportable by users; afterwards, these are reviewed with domain experts, with recurring failures treated as grounds for restricting generation in the affected area. The same loop covers bias, since patterns in which errors or omissions cluster around particular patient groups, languages, or contexts emerge only through review of accumulated cases.

Figure 5 presents MedCheck AI, a medication safety system that validates the combination of different medications for a patient. The mockup demonstrates DP 3.1 (Prioritise Domain-Grounded Plausibility) through clinical assessment grounded in pharmacological knowledge, explaining that combining these medications 'increases bleeding risk' and detailing the pharmacological mechanism by which they interact. DP 3.2 (Implement Criticality-Calibrated Validation) is shown through the 'Moderate' severity rating and comprehensive safety checks that include verification against DrugBank, the patient's current INR (2.3, within range), providing context-specific validation tailored to this patient's medical conditions. This illustrates criticality-calibrated validation at the high-stakes end: because a medication-interaction error is potentially severe, validation is correspondingly strict, combining structured-source verification, patient-specific grounding, and severity flagging that signals when human confirmation is required. A lower-stakes educational explanation, such as the CareGuide AI learning scenario (Figure 6), by contrast warrants lighter validation focused on domain accuracy rather than model-output verification.

### 4.4. Design principles for personalised adaptation (MR4)

#### 4.4.1. Design principle 4.1 (adapt dynamically to user characteristics)

For users differing in expertise, language proficiency, and cultural background (e.g. trainees vs. veterans, non-native speakers), the system should provide real-time user modelling that adapts vocabulary, conceptual depth, and language, to achieve personalised explanation matched to the individual user rather than a uniform default.

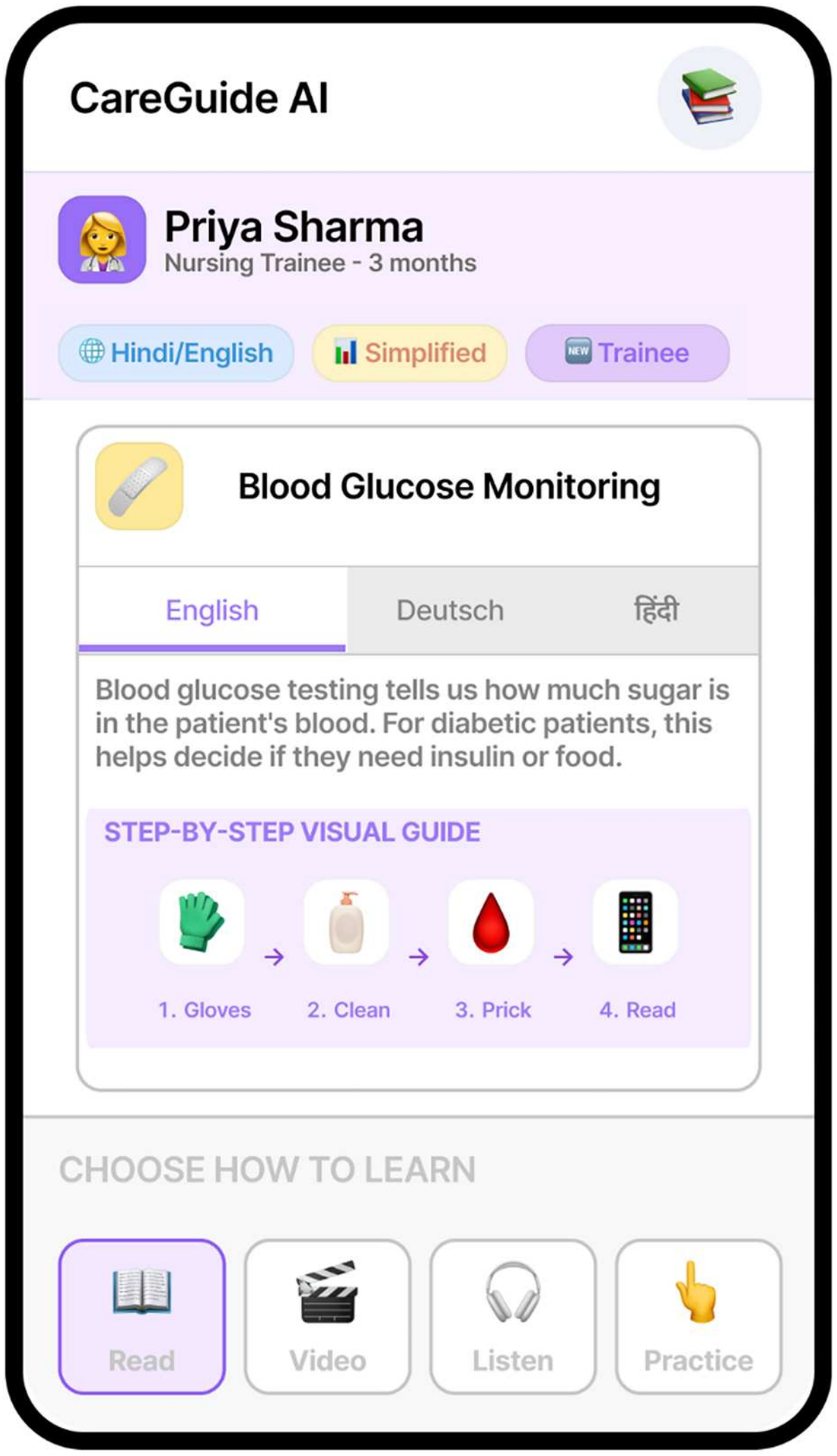

**Figure 6.** Mockup illustrating design principles 4.1 & 4.2 (CareGuide AI – care learning platform).

*4.4.1.1. Rationale.* Meta-Requirement 4 established need for personalised adaptation (Section 3.3.4). Users emphasised systems must 'adapt to the skill level of the caregiver' (U01-6) and 'level of experience' (U02-10), with linguistic diversity creating 'noticeable language barriers' (U06-4). Experts noted XAI personalisation is 'virtually non-existent' while generative AI enables adaptation from 'explain like I'm five' to technical formulas (AI-Exp-03).

*4.4.1.2. Implementation.* Implement user modelling capturing expertise levels, language proficiency, cultural background, and learning preferences through explicit settings or adaptive inference. Adjust vocabulary and conceptual depth based on user expertise. Enable multilingual explanation generation with culturally appropriate metaphors and examples. Consider daily routines and situational contexts affecting engagement. Adapt explanation detail based on time pressure and user focus.

#### 4.4.2. Design principle 4.2 (provide inclusive modality options)

For users with differing sensory preferences and accessibility needs, the system should provide multiple selectable presentation modalities (text, speech, visual) with accessibility accommodations, to achieve universal accessibility rather than a single fixed output format.

*4.4.2.1. Rationale.* Meta-Requirement 4 requires ensuring inclusive accessibility (Section 3.3.4). Users demonstrated varying preferences with some 'more visually oriented' while others 'rely on written text' (U03, U04), emphasising need for videos and multimodal formats (U02-9, U05-5). Experts highlighted benefits for users 'who are blind or have visual impairment' (AI-Exp-04), noting multimodal options address accessibility needs XAI cannot.

*4.4.2.2. Implementation.* Provide multiple presentation formats allowing user selection: text, spoken audio with adjustable speed, visual demonstrations, and interactive elements. Ensure screen reader compatibility and alternative text for visualisations. Support keyboard navigation. Offer adjustable text sizing and contrast modes. Enable users to control modality preferences explicitly. Design for universal accessibility as core capability rather than add-on feature.

Figure 6 shows CareGuide AI, an adaptive learning platform for healthcare education. The mockup features a nursing trainee learning about blood glucose monitoring. DP 4.1 (Adapt Dynamically to User Characteristics) is demonstrated through personalised settings that adjust content based on role (Nursing Trainee – 3 months), language preferences (Hindi/English), and complexity level (Simplified). The interface presents a step-by-step visual guide with numbered icons showing the glucose testing procedure. DP 4.2 (Provide Inclusive Modality Options) is shown through multiple learning formats available at the bottom: Read (book icon), Video (film icon), Listen (headphones icon), and Practice (hand icon), allowing learners to engage with content through their preferred modality.

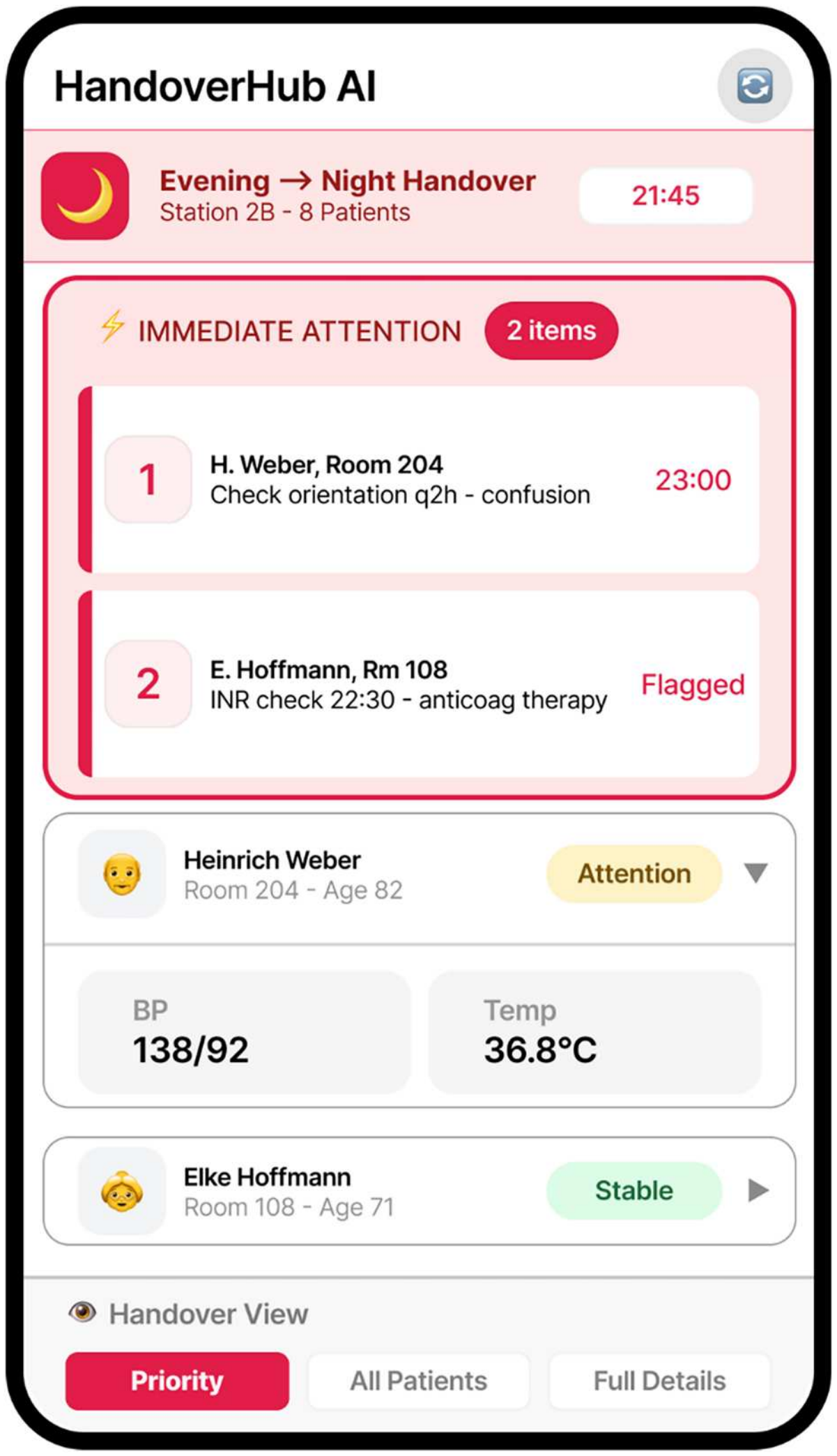

**Figure 7.** Mockup illustrating design principles 5.1 & 5.2 (HandoverHub AI – shift handover).

### 4.5. Design principles for cognitive load management (MR5)

#### 4.5.1. Design principle 5.1 (implement progressive information architecture)

For users under time pressure or cognitive fatigue (e.g. during busy shifts or after consecutive shifts), the system should provide a layered information architecture with progressive, user-controlled disclosure, to achieve minimised cognitive load without sacrificing access to depth.

*4.5.1.1. Rationale.* Meta-Requirement 5 established need for cognitive load management (Section 3.3.5). Users experiencing exhaustion described autopilot behaviour (U04-3) and being overwhelmed by information overload, with observations showing users 'often skip long texts and only read the first sentence' (U01). Experts described XAI as 'complete information overload' with '30 plots slapped down' creating overwhelming experiences (AI-Exp-02, AI-Exp-03). Users prefer 'being guided than being shown everything at once' (U01, U05).

*4.5.1.2. Implementation.* Design layered information structures presenting core information initially with progressive expansion options. Implement user-controlled depth mechanisms allowing expansion when desired rather than comprehensive simultaneous presentation. Create scannable initial layers supporting quick comprehension in time-pressured contexts. Break complex explanations into manageable chunks revealed sequentially or on-demand. Provide clear navigation between information layers. Design for situations requiring rapid understanding versus extended learning.

#### 4.5.2. Design principle 5.2 (filter and prioritise relevant information)

For users who need to act on an explanation rather than audit it, the system should provide relevance-based filtering that surfaces decision-critical factors by default and hides comprehensive detail behind expansion, to achieve resolution of the transparency-comprehension paradox (more information does not mean more understanding).

*4.5.2.1. Rationale.* Meta-Requirement 5 requires resolving transparency-comprehension paradox (Section 3.3.5). Evidence showed 'too much explanation leading to more confusion' (U04) with users skipping extensive information. Experts warned '30 plots slapped down' creates overwhelming overload (AI-Exp-02). Users need focused relevant information, not comprehensive exposure: ‘users prefer being guided than being shown everything at once’ (U01, U05).

*4.5.2.2. Implementation.* Prioritise information based on relevance to user goals and decision contexts rather than algorithmic importance measures. Filter less relevant details to focus attention on decision-critical factors. Avoid simultaneous presentation of confidence intervals, uncertainty measures, and comprehensive feature attributions unless explicitly requested. Design default presentations emphasising essential actionable information. Provide access to comprehensive details through expansion rather than default display. Recognise that clarity trumps transparency.

Figure 7 presents HandoverHub AI, a shift handover system for transitioning care between evening and night shifts. The mockup demonstrates DP 5.1 (Implement Progressive Information Architecture) through hierarchical information display that prioritises urgent matters at the top, an ‘Immediate Attention’ section with 2 items highlights critical issues. Below this, patient cards show essential vitals. DP 5.2 (Filter and Prioritise Relevant Information) is evidenced by user-controllable view options at the bottom (Priority/All Patients/Full Details) and the attention-based categorisation system that allows staff to quickly focus on patients requiring immediate action while keeping stable patients visible but de-emphasised.

## 5. Discussion

### 5.1. From explainable to explanatory AI: design principles for human understanding

Our findings speak directly to the gap identified in Section 1: the absence of systematic, empirically grounded design knowledge for AI systems whose purpose is human understanding rather than model transparency; and to the two research questions derived from it. In answering what Explanatory AI is and which dimensions distinguish it from traditional XAI (RQ1), the five dimensions and their meta-requirements characterise the paradigm; in answering what meta-requirements and design principles should guide the development of Explanatory AI systems for high-stakes professional contexts (RQ2), the ten design principles translate that characterisation into actionable design knowledge.

This research establishes Explanatory AI as an emerging, design-oriented paradigm where AI systems leverage generative capabilities to serve as explanatory partners, grounded in theory and exploratory qualitative evidence. The ten design principles presented in Section 4 operationalise this emerging paradigm, providing actionable guidance for creating systems characterised by interpretive sense-making, dynamic narrative interaction, contextual plausibility, personalised adaptation, and cognitively-aligned information delivery. These principles represent a fundamental shift from model-centric transparency to user-centric communication.

#### 5.1.1. The narrative turn in AI explanation

The emergence of generative AI creates unprecedented opportunities for narrative-driven explanations that adapt to user contexts and needs (Chen et al. 2024; Martens et al. 2025). While traditional XAI methods present static visualisations or feature importance scores (Lundberg and Lee 2017; Ribeiro, Singh, and Guestrin 2016), Explanatory AI systems generate dynamic explanations incorporating analogies, cross-domain knowledge, and follow-up responses that mirror human explanatory conversations. This narrative approach presents both opportunities and challenges, requiring careful design to ensure explanations maintain appropriate complexity while remaining accessible (Martens et al. 2025). The communicative power of narrative explanations brings responsibility to inform rather than manipulate users through persuasive techniques. Moreover, the very accessibility that makes narrative explanations effective can become a liability: highly comprehensible explanations may lead users to infer information beyond what is actually conveyed, with users often remaining overconfident in these misinterpretations (Papenkordt, Ngonga Ngomo, and Thommes 2026; Xuan et al. 2025). This underscores the need for explanations to explicitly communicate not only what they convey but also their boundaries and limitations.

#### 5.1.2. Principles as architectural layers, not silos

While we present five dimensions and ten corresponding design principles separately for analytical clarity, we do not intend them as independent, non-overlapping requirements. Dynamic narrative interaction (MR2), personalised adaptation (MR4), and progressive information architecture (MR5) all govern how explanation content is delivered and necessarily interact: personalising an explanation’s complexity level inherently affects its narrative structure and information layering, and vice versa. We therefore conceptualise the five dimensions as interdependent layers within a single explanatory architecture, where design decisions at one layer constrain or enable decisions at others, rather than as a checklist of independent requirements to be satisfied in isolation. Future implementation work should examine how these layers interact in practice and whether certain combinations (e.g. high personalisation with minimal progressive disclosure) produce tensions requiring explicit design trade-offs.

#### 5.1.3. Trust through contextual communication

Our healthcare study provides compelling evidence that users' explanatory needs align with Explanatory AI principles. Caregivers sought explanations of clinical reasoning and domain knowledge, why certain care practices make sense, rather than algorithmic decision processes. Across all five dimensions, participants preferred systems serving as explanatory partners for domain understanding. Caregivers consistently valued narrative-driven, multimodal, cognitively manageable explanations adaptive to their specific contexts and capabilities. This finding suggests that trust emerges from meaningful communication about contextual reasoning rather than technical comprehension of algorithmic processes (Bove et al. 2024; Miller 2019; Shin 2021). Beyond communication content, how the AI system is framed, including its perceived power status and degree of anthropomorphisation, also influences trust, with higher-status framing increasing users' confidence in system outputs (Ha et al. 2022). This raises the question of whether a sufficiently persuasive or conversational XAI system becomes Explanatory AI. On our functional account, systems do not gradually transform into Explanatory AI along a single axis. Rather, Explanatory AI is defined by whether a system meets the meta-requirements and design principles that make human understanding its primary objective. Persuasiveness alone is not the criterion; in fact, persuasive explanation without evidential grounding is precisely what our epistemic-stance requirement rules out (Section 3.3.3). A conversational or human-centered XAI system that is designed to satisfy these meta-requirements is best understood as an instantiation of Explanatory AI, whereas one that makes model outputs more engaging without prioritising contextual understanding, domain grounding, and cognitively aligned delivery remains within the XAI paradigm. The distinction is thus not one of degree along a persuasiveness scale, but of design intent and satisfaction of the requirements we specify.

#### 5.1.4. Inclusivity and the democratisation of AI understanding

The emphasis on inclusivity and personalisation proves especially critical in domains marked by linguistic diversity and varying technical expertise levels. Current one-size-fits-all XAI approaches systematically exclude non-expert users, creating barriers precisely where explainability matters most (Kim et al. 2024; Meske et al. 2022). Explanatory AI's adaptive capabilities offer pathways to more inclusive AI systems serving diverse user populations effectively (Singh et al. 2024; Trewin et al. 2019). Our model extends accessibility considerations by incorporating real-time adaptation to user characteristics, situational context, and cultural background as core system capabilities rather than add-on features.

#### 5.1.5. Technical implementation challenges and considerations

While our conceptual model demonstrates Explanatory AI's potential, several technical challenges must be addressed for successful implementation. Recent advances in generative AI provide the technical foundation for narrative explanation systems (Vaswani et al. 2017; Wei et al. 2022), but real-world applications require addressing challenges around data quality, computational requirements, and ethical considerations. Key considerations include: ensuring narrative explanations maintain sufficient accuracy while remaining comprehensible, developing real-time personalisation algorithms, creating evaluation frameworks balancing technical fidelity with communicative effectiveness, and addressing computational costs of generating contextual, personalised explanations at scale. More broadly, integrating LLMs into socio-technical systems raises questions around governance, accountability, and the division of cognitive labour between humans and AI that extend beyond technical performance metrics (Torkamaan et al. 2024). Empirical studies of generative AI integration in knowledge work further highlight that effective human-AI partnerships require workers to develop new skills in critiquing, integrating, and communicating AI-generated outputs (Jarrahi et al. 2025). The risk of hallucinated or oversimplified explanations represents particular concern, requiring sophisticated quality control mechanisms and ongoing evaluation (Singh et al. 2024; Zhang et al. 2025).

#### 5.1.6. Rethinking the detail-comprehensibility trade-off

Our model challenges the assumption that exhaustive explanation detail must always take precedence over comprehensibility. Explanatory AI proposes optimising for contextual plausibility and actionable insight while maintaining sufficient accuracy to support sound decision-making (Miller 2019). This rebalancing concerns comprehensibility, not correctness: plausibility does not substitute for fidelity, truthfulness, or domain validity, and in high-stakes contexts these remain non-negotiable regardless of how comprehensible an explanation is. The trade-off is thus not between accuracy and comprehensibility per se, but about how much model-level detail to surface, bounded by the requirement that what is surfaced remain faithful to its reference object and domain-valid, and by the distinction between reduction of detail and misrepresentation set out in Section 3.3.3. This approach requires careful consideration of contexts where strict fidelity is essential versus those where comprehensible approximation serves users better. High-stakes regulatory environments may require hybrid approaches providing both technical records for audit purposes and communicative clarity for human understanding, recognising these serve different explanatory purposes (Meske et al. 2022).

#### 5.1.7. Design principle interactions and trade-offs

The ten design principles are not independent switches to be maximised in isolation; in a deployed Explanatory AI system they interact, and pursuing one can work against another. Two examples illustrate this. First, aggressive personalisation (DP 4.1) can pull against domain-grounded plausibility and validation (DP 3.1, DP 3.2): simplifying an explanation for a novice may omit qualifications that matter for correctness, raising the risk of confident but incomplete understanding precisely for the users least able to detect it. Second, progressive disclosure and relevance-based filtering (DP 5.1, DP 5.2) can conflict with the completeness that criticality-calibrated validation may demand: information withheld to manage cognitive load in a high-stakes decision could be exactly the caveat a practitioner needs. These tensions do not have a single correct resolution, because the appropriate balance depends on user expertise, task stakes, and domain. Rather than prescribing fixed weightings, we therefore recommend that designers treat the principles as a system to be balanced deliberately: making trade-offs explicit for the target context, attending to interactions that affect safety, and calibrating the balance to criticality rather than applying each principle uniformly. Systematically characterising these interactions, and empirically evaluating how particular combinations affect comprehension, trust calibration, and decision quality, is an important direction for future work. This includes determining how validation intensity should scale with criticality, and establishing the domain-specific acceptance thresholds that we deliberately leave open here.

### 5.2. Implications for research and practice

The emergence of Explanatory AI has profound implications for AI system design and organisational implementation. Instead of treating explainability as a post-hoc addition to algorithmic systems, developers must consider explanatory effectiveness as a core component of system performance when the goal is human understanding (Singh et al. 2024). This is consistent with the Human-Centred AI paradigm, in which augmenting human understanding, rather than replacing human judgment, is a primary purpose, alongside ensuring appropriate oversight and integrity (Schmager, Pappas, and Vassilakopoulou 2025). The ability to generate contextually appropriate, culturally sensitive explanations becomes as important as underlying decision-making capabilities when systems serve as explanatory partners (Chen et al. 2024). Research on generative AI in knowledge work confirms this dynamic human-AI partnership, where workers actively contribute critique, contextualisation, and individualisation of AI outputs rather than passively consuming them (Jarrahi et al. 2025). This approach requires interdisciplinary collaboration between AI researchers, UX designers, domain experts, and end users, as types of explanations and their perceiving varies greatly (Miller 2019). Building on the multi-stakeholder perspective (Meske et al. 2022), traditional AI development processes that prioritise algorithmic performance metrics must expand to include measures of communicative effectiveness, user comprehension, and practical utility in real-world contexts when systems are designed to support human understanding. Organisations implementing Explanatory AI systems must invest in user research, cultural competency, and iterative design processes that prioritise human understanding over technical elegance. This necessitates new training programmes for staff who must learn to work with AI systems that communicate differently than traditional decision support tools, developing skills in AI-mediated dialogue while maintaining critical thinking when working with systems capable of persuasive communication (Bove et al. 2024). Additionally, organisations must account for how explanation design influences not only task performance but also psychological closeness and identity threat (Wang, Zhuang, and Santoso 2026). Such implementation requires theory-driven, user-centric design approaches that prioritise human needs over technical convenience (Wang et al. 2019).

As regulatory frameworks increasingly require meaningful explanations of automated decision-making, the distinction between explainable AI and Explanatory AI becomes practically important. Traditional XAI approaches often produce explanations that satisfy audit requirements but fail to help affected individuals understand AI decisions (Bove et al. 2024; Chromik and Schuessler 2020). Users differ in whether they prefer system transparency through explainability features or rely more on external trust signals such as AI certification seals, raising design challenges when both groups must be served (Cetinkaya and Krämer 2026). Organisations may need dual-purpose systems: explainable AI components that provide technical transparency for regulatory compliance, and Explanatory AI components that generate meaningful explanations for human comprehension. Whether to deploy XAI, Explanatory AI, or both depends on the purpose the explanation must serve rather than on a ranking of the two paradigms. XAI is the appropriate choice when the goal is model validation, debugging, auditing, or regulatory compliance, that is, when a stakeholder must verify how a model arrived at an output. Explanatory AI is appropriate when the goal is human comprehension, informed decision-making, or learning, that is, when a user must understand why a recommendation makes sense in context. Many real settings require both at once: a clinical decision-support system may need XAI to satisfy audit and accountability requirements while offering Explanatory AI to help clinicians reason about and act on its outputs. In such cases the two operate on different targets, the model on one side and the user's understanding on the other, and are therefore complementary rather than substitutive. Balancing them is thus less a matter of choosing between paradigms than of allocating each to the purpose it serves and calibrating their relative prominence to the stakes and the audience: higher-criticality or regulated contexts warrant stronger XAI-based verification alongside Explanatory AI, whereas lower-stakes or educational contexts may rely primarily on Explanatory AI. This separation acknowledges that regulatory auditing and human understanding serve different purposes and may require different approaches. However, this dual approach creates new challenges for validation and quality assurance, requiring organisations to develop governance frameworks ensuring both technical accuracy for compliance purposes and communicative effectiveness for human understanding. Our findings open opportunities for complementary deployment where explainable AI serves validation purposes while Explanatory AI serves end users, with each system optimised for its specific explanatory purpose rather than attempting to serve all stakeholders with a single approach.

The shift towards Explanatory AI requires fundamental changes in how organisations approach AI implementation and user training. Traditional approaches that focus on algorithmic performance and technical validation must expand to include user experience design, communication effectiveness, and cultural sensitivity. The trajectory towards more interactive AI systems that provide users with greater agency beyond mere explanation or contestability (Raees et al. 2024) suggests that Explanatory AI represents one component of a broader shift towards genuinely collaborative human-AI partnerships. Organisations must develop new competencies in narrative design, cross-cultural communication, and adaptive user interface development. This includes establishing new roles and responsibilities, such as explanation designers who specialise in creating effective AI-human communication, and cultural liaisons who ensure systems work effectively across diverse user populations. The healthcare context of our study illustrates these challenges particularly clearly, where explanatory systems must work effectively across multiple languages, varying levels of technical expertise, and high-stress environments where cognitive load management is critical.

Implementation of Explanatory AI systems also requires new approaches to quality assurance and system validation. While XAI evaluation has progressed towards human-centered metrics (Hoffman et al. 2018), Explanatory AI requires that systems must be evaluated based on their effectiveness in supporting human understanding and decision-making. This requires developing new testing methodologies that assess comprehension, user satisfaction, and practical utility in real-world contexts. Organisations must establish feedback mechanisms that capture user experiences with AI explanations and iteratively improve system performance based on human-centered criteria rather than purely technical metrics.

### 5.3. Limitations

Several limitations constrain the generalizability and applicability of our findings. First, the empirical evidence derives from a single domain (home care in Germany) with specific characteristics including diverse user populations with varying language proficiency, high-stakes care decisions, and time-pressured work environments. While we argue these characteristics make design needs particularly visible and reveal requirements relevant across domains where non-technical users must understand and act on AI recommendations, systematic validation across additional contexts remains necessary. Healthcare, educational, financial, and legal domains each present unique constraints and user characteristics that may reveal additional design requirements or modify the relative importance of specific principles. Relatedly, our characterisation of Explanatory AI as an emerging paradigm rests on exploratory qualitative evidence, and broader validation is needed before its articulation as a design paradigm can be considered established rather than proposed. Generalizability can be strengthened along a clear path: replicating the derivation in contrasting high-stakes domains (e.g. financial or legal decision support) to test which meta-requirements are domain-invariant and which require adaptation and complementing the qualitative foundation with larger-scale studies. We position home care not as a representative sample but as a revelatory case whose diversity of user expertise, language proficiency, and time pressure surfaces explanation needs that are likely present, though less visible, in other domains where non-experts must act on AI recommendations.

Second, our study employed qualitative methods with limited sample sizes. Contextual interviews with six nursing staff members and six AI experts provide rich insights into user needs and expert perspectives but constrain statistical generalisation. The sample sizes were appropriate for theory-informed design and principle derivation but insufficient for hypothesis testing or quantitative validation of principle effectiveness. Future research requires larger-scale studies measuring the impact of Explanatory AI systems on comprehension, decision quality, trust calibration, and learning outcomes across diverse user populations. Our empirical analysis also applied the five dimensions deductively. While bottom-up clustering and a residual category allowed material outside the framework to be retained and inspected, a deductive frame is inherently better suited to elaborating existing dimensions than to discovering new ones, and inductive or larger-scale analysis could yield dimensions our framework does not anticipate.

Third, we have not implemented or evaluated actual systems embodying these principles. Relatedly, the UI mockups presented in Section 4 are illustrative design artefacts created by the authors to visualise plausible implementations within our study's domain. They have not undergone user testing or expert validation and should not be read as evidence of design principle effectiveness. Future validation of these or comparable mockups should involve structured expert review (e.g. domain experts and HCI researchers assessing principle-feature correspondence) followed by user evaluation with target populations such as nursing staff, assessing comprehension, usability, and decision support quality. The principles remain theoretical constructs derived from integrating multidisciplinary theory with empirical requirements, requiring technical implementation and user evaluation to demonstrate feasibility and effectiveness. Future work should address technical challenges including real-time adaptation mechanisms, hallucination detection and mitigation, computational efficiency at scale, and integration with existing organisational systems. System evaluation should examine whether implementations genuinely improve user comprehension and decision-making compared to traditional XAI approaches or baseline explanations. We are therefore explicit that this paper does not provide evidence that Explanatory AI improves comprehension, trust calibration, decision quality, or safety; establishing such effects requires the controlled evaluations we outline above and defer to future work. What the contribution does provide is prescriptive design knowledge derived systematically by triangulating multidisciplinary explanation theory with empirical evidence from end users and technical experts. Following established design theory (Gregor, Chandra Kruse, and Seidel 2020), such theory-informed design knowledge is a legitimate and necessary contribution in its own right: it makes the design space explicit, renders the underlying rationale inspectable and contestable, and provides the testable propositions that subsequent empirical work can confirm or refute. Rather than asserting effectiveness, the framework's value lies in structuring how such effectiveness can be investigated, and in offering practitioners actionable guidance that is grounded in theory and user needs rather than intuition.

Fourth, our literature synthesis followed a purposive, integrative-review approach (Torraco 2005) rather than a systematic review with a fixed search protocol; while appropriate for our goal of conceptual framework-building across heterogeneous literatures, it does not guarantee exhaustive coverage of relevant work, and a systematic or scoping review could surface additional theoretical perspectives or boundary cases not captured here. Relatedly, our account of human explanation identifies principles that recur across the literatures we surveyed rather than a complete taxonomy of human explanation types, and further principles may become relevant as the framework is applied in other domains.

## 6. Conclusion

This paper introduces Explanatory AI as an emerging, design-oriented paradigm complementary to explainable AI, where AI systems leverage generative capabilities as explanatory partners for human understanding. Through theory-informed design integrating multidisciplinary explanation theory with empirical evidence from healthcare contexts and AI expert interviews, we established a framework identifying five key dimensions: explanatory purpose, communication mode, epistemic stance, adaptivity, and cognitive design. We derived five meta-requirements specifying what Explanatory AI systems must achieve and formulated ten design principles prescribing how to build such systems. Our findings demonstrate that users require explanations fundamentally different from traditional XAI outputs, seeking interpretive sense-making, narrative communication, contextual plausibility, personalised adaptation, and cognitively-aligned delivery. Explanatory AI addresses these needs while complementing XAI's essential role in model validation and regulatory compliance. Systems can operate standalone or in combination with XAI, with each serving distinct purposes in the AI explanation ecosystem.

Future research should address several critical directions. First, systematic validation across diverse domains beyond healthcare is needed to assess generalizability. Second, implementation studies must demonstrate technical feasibility and evaluate system effectiveness through controlled user studies. Third, research must develop robust validation mechanisms addressing hallucination risks, particularly for high-stakes domains. Fourth, investigation of hybrid architectures combining XAI and Explanatory AI components would illuminate optimal deployment strategies. Finally, longitudinal studies examining effects on user learning, trust calibration, and decision quality would provide crucial insights. We therefore present Explanatory AI as an emerging, theoretically grounded and empirically informed design direction: the meta-requirements and design principles offer a systematic foundation whose effectiveness remains to be established through prototyping, implementation, and controlled evaluation. Positioned this way, the framework marks a starting point for research on AI systems whose purpose is human understanding. By shifting focus from algorithmic transparency to human understanding, Explanatory AI offers pathways towards AI systems genuinely supporting human reasoning across diverse contexts.

## Author contributions

CRediT: **Christian Meske:** Conceptualization, Funding acquisition, Investigation, Methodology, Project administration, Resources, Writing – original draft, Writing – review & editing; **Justin Brenne:** Conceptualization, Data curation, Investigation, Methodology, Writing – original draft, Writing – review & editing; **Erdi Ünal:** Conceptualization, Investigation, Methodology, Writing – original draft, Writing – review & editing; **Sabahat Ölcer:** Data curation, Investigation, Methodology, Writing – review & editing; **Aysegül Dogangün:** Data curation, Investigation, Methodology, Writing – review & editing.

## Disclosure statement

No potential conflict of interest was reported by the author(s).

During the preparation of this work the authors used OpenAI's ChatGPT 5.3 as well as Anthropic's Claude Sonnet 4.8 to improve language clarity, grammar and style. After using these tools/services, the authors reviewed and edited the content as needed and take full responsibility for it.

## Funding

The authors would like to express their gratitude to all nurses and AI experts who participated in the interviews and generously shared their knowledge and experiences. This work was supported by the German Federal Ministry of Research, Technology and Space (BMFTR) under the project "ExpliCare-Next" (grant number 01MK24003F).